\documentclass[]{interact}
\usepackage[caption=false]{subfig}
\usepackage[numbers, sort&compress]{natbib} 
\usepackage{booktabs} 
\bibpunct[, ]{[}{]}{,}{n}{,}{,}

\makeatletter
\def\NAT@def@citea{\def\@citea{\NAT@separator}}
\makeatother
\usepackage[tablesfirst, notablist, figlist, disable]{endfloat} 

\usepackage[colorlinks=true, allcolors=blue]{hyperref}
\usepackage{url}
\usepackage{siunitx}
\usepackage{mathtools}
\usepackage{multirow}
\usepackage{colortbl}
\usepackage{appendix} 
\usepackage{xcolor}
\usepackage[english]{babel}
\usepackage{placeins}
\newcommand{\eqdot}{\, .} 
\newcommand{\eqcomma}{\, ,} 
\newcommand{\LU}[2]{\prescript{#1}{}{#2}\,}
\newcommand{\LUR}[3]{\prescript{#1}{}{#2}_{#3}\,}

\newcommand{\vp}[2]{\left[ \begin{array}{c} { #1} \vspace{0.1cm}\\ { #2} \end{array} \right]}
\newcommand{\vr}[3]{\left[ \begin{array}{c} { #1}\vspace{0.1cm} \\ { #2}\vspace{0.1cm} \\ { #3} \end{array} \right]}
\newcommand{\vtretp}[3]{\left[ \begin{array}{ccc} #1 & #2 & #3 \end{array} \right]}
\renewcommand{\mp}[4]{\left[ \begin{array}{cc} #1 & #2 \vspace{0.1cm}\\ #3 & #4 
\end{array} \right]}
\newcommand{\av}{\mathbf{a}}
\newcommand{\bv}{\mathbf{b}}
\newcommand{\cv}{\mathbf{c}}
\newcommand{\ev}{\mathbf{e}}
\newcommand{\fv}{\mathbf{f}}
\newcommand{\hv}{\mathbf{h}}
\newcommand{\qv}{\mathbf{q}}
\newcommand{\rv}{\mathbf{r}}
\newcommand{\sv}{\mathbf{s}}
\newcommand{\tv}{\mathbf{t}}
\newcommand{\vv}{\mathbf{v}}
\newcommand{\xv}{\mathbf{x}}
\newcommand{\yv}{\mathbf{y}}
\newcommand{\zv}{\mathbf{z}}
\newcommand{\Rm}{\mathbf{R}}
\newcommand{\Tm}{\mathbf{T}}
\newcommand{\Kframe}{\ensuremath{\mathrm{K}}}
\newcommand{\Mframe}{\ensuremath{\mathrm{M}}}
\newcommand{\tomega}{\mbox{$\pmb{\omega}$}\!\!}
\newcommand{\tp}{^\mathrm{T}}
\newcommand{\Null}{\mathbf{0}}
\newcommand{\One}{\mathbf{1}}
\newcommand{\Rotation}[2]{\ensuremath{\mathrm{Rot_#1\left(#2\right)}}}
\newcommand{\angularVelocity}[2]{\ensuremath{\LUR{#1}{\tomega}{~ #2}}}
\newcommand{\angularVelocitySkew}[2]{\ensuremath{\LUR{#1}{\,\widetilde{\tomega}}{~ #2}}}
\newcommand{\Gframe}{\ensuremath{\mathrm{0}}}
\newcommand{\Rframe}{\ensuremath{\mathrm{R}}}
\newcommand{\Fframe}{\ensuremath{\mathrm{F}}}
\newcommand{\Bframe}{\ensuremath{\mathrm{B}}}
\newcommand{\Hframe}{\ensuremath{\mathrm{H}}}
\newcommand{\BIframe}{\ensuremath{\mathrm{B_1}}}
\newcommand{\BIIframe}{\ensuremath{\mathrm{B_2}}}
\newcommand{\Bsframe}{\ensuremath{\mathrm{B^*}}}
\newcommand{\HIIframe}{\ensuremath{\mathrm{H_2}}}
\newcommand{\HIIIframe}{\ensuremath{\mathrm{H_3}}}
\newcommand{\Hsframe}{\ensuremath{\mathrm{H^*}}}
\newcommand{\Cframe}{\ensuremath{\mathrm{C}}}
\newcommand{\TIframe}{\ensuremath{\mathrm{T_1}}}
\newcommand{\TIIIframe}{\ensuremath{\mathrm{T_3}}}
\newcommand{\Pp}{\ensuremath{\mathrm{P}}}
\newcommand{\Qp}{\ensuremath{\mathrm{Q}}}
\newcommand{\PI}{\ensuremath{\mathrm{P_1}}}
\newcommand{\PII}{\ensuremath{\mathrm{P_2}}}
\newcommand{\PIII}{\ensuremath{\mathrm{P_3}}}
\newcommand{\rR}{\ensuremath{\mathrm{r_R}}}
\newcommand{\rF}{\ensuremath{\mathrm{r_F}}}
\newcommand{\dI}{\ensuremath{\mathrm{d_1}}}
\newcommand{\dII}{\ensuremath{\mathrm{d_2}}}
\newcommand{\dIII}{\ensuremath{\mathrm{d_3}}}
\newcommand{\alp}{\ensuremath{\mathrm{\alpha}}}
\newcommand{\lam}{\ensuremath{\mathrm{\lambda}}}
\newcommand{\BCOM}{\ensuremath{\mathrm{B_{COM}}}}
\newcommand{\HCOM}{\ensuremath{\mathrm{H_{COM}}}}
\newcommand{\HT}[2]{\ensuremath{\mp{#1}{#2}{\Null}{1}}}
\newcommand{\EXU}{Exudyn}
\newcommand{\SB}{Stable-Baselines3}
\newcommand{\steps}[1]{\ensuremath{n_\mathrm{#1}}}
\begin{document}
\title{Path Following and Stabilisation of a Bicycle Model using a Reinforcement Learning Approach}
\author{\name{Sebastian Weyrer\textsuperscript{a}, Peter Manzl\textsuperscript{a}, A. L. Schwab\textsuperscript{b}, Johannes Gerstmayr\textsuperscript{a$\ast$}\thanks{$^\ast$ Corresponding author. E-mail: johannes.gerstmayr@uibk.ac.at}}
\affil{\textsuperscript{a}Department of Mechatronics, University of Innsbruck, Innsbruck, Austria; \textsuperscript{b}Department of Biomechanical Engineering, Delft University of Technology, Delft, The Netherlands}}
\maketitle

\begin{abstract}
Over the years, complex control approaches have been developed to control the motion of a bicycle. Reinforcement Learning (RL), a branch of machine learning, promises easy deployment of so-called agents. Deployed agents are increasingly considered as an alternative to controllers for mechanical systems. The present work introduces an RL approach to do path following with a virtual bicycle model while simultaneously stabilising it laterally. The bicycle, modelled as the Whipple benchmark model and using multibody system dynamics, has no stabilisation aids. The agent succeeds in both path following and stabilisation of the bicycle model exclusively by outputting steering angles, which are converted into steering torques via a PD controller. Curriculum learning is applied as a state-of-the-art training strategy. Different settings for the implemented RL framework are investigated and compared to each other. The performance of the deployed agents is evaluated using different types of paths and measurements. The ability of the deployed agents to do path following and stabilisation of the bicycle model travelling between \SI{2}{ms^{-1}} and \SI{7}{ms^{-1}} along complex paths including full circles, slalom manoeuvres, and lane changes is demonstrated. Explanatory methods for machine learning are used to analyse the functionality of a deployed agent and link the introduced RL approach with research in the field of bicycle dynamics.
\end{abstract}

\section{Introduction}
\label{Introduction}
Bicycle dynamics is a discipline within dynamics studying the motion of bicycles under the influence of forces. Understanding bicycle dynamics is essential for developing controllers for bicycle models.

Traditionally, controllers for mechanical systems like bicycles are designed using principles from control engineering. Alongside the approaches from control engineering, there is an emerging interest in alternative approaches using machine learning methods to control mechanical systems. There is an increasing trend in multibody systems to apply machine learning methods, in the range between feed-forward networks for surrogate models \cite{Choi2021} and large language models to model dynamic systems \cite{Gerstmayr2024}. In particular, Reinforcement Learning (RL), a sub-area of machine learning, has been successfully used to solve control tasks with various multibody systems \cite{Hashemi2023}. The idea of RL is that an agent learns to solve a task by observing the dynamic environment, receiving a reward signal, and returning actions to the dynamic environment without being supervised \cite{Zai2020}.

This raises the question of whether RL could serve as a viable approach for controlling bicycle models, similar to the controllers designed using traditional control engineering methods. A possible application of the RL approach alongside conventional control engineering approaches for controlling bicycles is not only of theoretical interest but could also be in practical robotic systems in the future. Against the background of this question, the state-of-the-art is considered from two perspectives. Firstly, research on the control of the motion of a bicycle, focusing on control engineering methods, is presented below. Secondly, existing work on using RL and related approaches with bicycle models is considered.

The control of the motion of a bicycle has attracted many researchers over the years \cite{Meijaard2007}. Probably because of the speed-dependent lateral stability and the non-minimal phase behaviour, that is, to stabilise a bicycle one has to steer into the direction of the undesired fall \cite{Kooijman2011}. Most of the attention has been directed towards stabilising the lateral dynamics of the bicycle and not so much on path following. A good overview on the dynamics and control of a bicycle is presented in \cite{Schwab2013}. After 2013, more work emerged on the control of the bicycle. The focus of the present work is on path tracking, which of course must include lateral stability control. A passive velocity field controller for path following has been introduced by Yin and Yamakita \cite{Yin2016}. Their system is then passive with respect to external force perturbations. Turnwald et al. \cite{Turnwald2018} introduce a passivity-based path following control for a bicycle. The steering and forward speed are controlled using a generalised canonical transformation on their port-Hamiltonian system of the bicycle. A classic multi-loop control structure is used by Shafiei and Emami \cite{Shafiei2019} for path following of an unmanned bicycle. A classic LQR based controller for straight path following and cornering for a bicycle robot has been designed and built by Seekhao et al. \cite{Seekhao2020}, where the lateral dynamics of the bicycle is stabilised by an additional non-inverted pendulum. A model predictive controller with constraints on the lean, steer, heading as well as position of the bicycle is used by Persson et al. \cite{Persson2021Trajectory} for the design of a path following controller for a riderless bicycle. He et al. \cite{He2022} present a learning based control framework for path following of a bicycle robot. For path following both steer and forward speed actuation are used. The lateral dynamics of the bicycle robot is stabilised by a controlled non-inverted pendulum.

The idea of using an RL approach to control bicycle models is not new. An article from Randlov and Alstrøm \cite{Randlov1998} published in 1998 shows that RL can be used to keep a bicycle model upright and steer it to a target point that is \SI{1000}{m} away from the initial position of the vehicle. The paper suggests using shaping, where two separate neural networks are used. One net keeps balance and the other steers the bicycle towards the target point. The path the bicycle can take to the target point is not predefined. A good 20 years later, the articles from Le and Chung \cite{Tuyen2017} and Le et al. \cite{Tuyen2019} show that shaping is not necessary when further evolved RL algorithms are used, as a single agent learns to steer the bicycle model to the target point without it falling over. The latter two works mainly adopt the bicycle model from the work \cite{Randlov1998}. This bicycle model is characterised by not having a trail. The trail is a geometric property of a bicycle, substantially contributing to the speed-dependent lateral stability of the bicycle \cite{Meijaard2007}. In an approach that does not use RL per se, but shows that neural networks, in general, can be useful for controlling bicycles, Cook \cite{Cook2004} manually devised a neural network controller with only two neurons as an example of a simple human bicycle tracking controller. The first neuron is a proportional controller on the heading with a threshold function. It outputs a desired roll angle, which is an input for the second neuron which in turn outputs a steering torque based on PD control. The desired heading is set using waypoints enabling the bicycle to perform tracking tasks. He finds that the controller is relatively robust as gain values weakly depend on speed and do not have to be perfectly adjusted to the specific bicycle. The controller works at a range of velocities, but it fails at low speed. A work from Zhu et al. \cite{Zhu2021} uses an RL approach to steer a bicycle model along a predefined path, but does not consider lateral stabilisation of it. The therein used bicycle model has an integrated inertia wheel to enhance its stability. Taken together, the referenced works do not show that the RL approach can be used both to stabilise the bicycle model and to do path following, as they specialise in either one of these tasks. Moreover, the aforementioned works on using RL to stabilise a bicycle model do not clarify whether RL can induce a speed-dependent lateral stabilisation policy, as required for riding the positive trail bicycle we ride every day.

The objective of the present work is to show for the first time that an RL approach can be used to do path following with a bicycle model while simultaneously stabilising it laterally. In the present work, the bicycle model has no balance aids such as inertia wheels or non-inverted pendulums. The agent must learn to do path following and stabilisation of the bicycle model exclusively by turning the handlebar of the bicycle. The bicycle in the present work is modelled using the so-called Whipple bicycle \cite{Whipple1899}, which is regarded as a benchmark model to study bicycle dynamics \cite{Xiong2020}. The Whipple bicycle reflects the dynamic characteristics of the bicycle found on the road today, having a positive trail and being unstable when moving at low forward velocities and self-stable when moving at a specific speed range \cite{Meijaard2007}. This brings another novelty to the present work, namely the investigation of whether an RL approach is able to induce a speed-dependent policy to stabilise a bicycle model having a positive trail. A further feature of the present work is that the agent is not allowed to change the forward velocity of the bicycle model by itself but must be able to cope with different forward velocities, underlining the practical significance of the present work by addressing a system in which the velocity of the bicycle is set by the cyclist while the agent does path following and stabilisation of the bicycle model by doing steering movements. Another specification of the present work is that the trained agent must be able to steer the bicycle model along arbitrary paths containing full circles, polynomials, straight lines, and curves. By using explanatory methods for machine learning, the present work aims to verify that the agents learn the basic mechanisms of controlling a bicycle. 

An additional aspect of the present work is that the environment is virtual, as such that the bicycle is implemented as a multibody model. Using a virtual environment is accompanied by simplifications resulting from the bicycle model and its surrounding. It is assumed that the bicycle model drives on a perfectly levelled surface.

The simplifications and mandatory preparatory work associated with the bicycle model to apply an RL approach mark the beginning of this article. Afterwards, the applied RL framework is described, followed by an outline of the learning process with the application of curriculum learning as a training strategy. The subsequent section evaluates how accurately the agents can do path following and stabilisation of the bicycle model at different forward velocities and along two different types of paths. Explanatory methods for machine learning are used to analyse the functioning of the agents. Finally, the presented results are discussed.

\section{Bicycle Model and Associated Preparatory Work}
\label{Bicycle Model and Associated Preparatory Work}
The bicycle model used in the present work is the Whipple bicycle \cite{Whipple1899}. The Whipple bicycle is often used to study bicycle dynamics \cite{Xiong2020} and the linearised equations of motion for the Whipple bicycle are presented in a benchmark paper \cite{Meijaard2007}. The linearised equations of motion are confirmed by comparing the eigenvalues of the system predicted by the linearised equations of motion with experimentally measured eigenvalues of a real bicycle \cite{Kooijman2008}. By using the Whipple bicycle, the RL approach shown in the present work is applied to a system of which the control task changes with the forward velocity. In this section, the bicycle model itself and the preparatory work that is required to use the bicycle model with the RL approach are described.

\begin{figure}[b]
	\centering
	\includegraphics[scale=1]{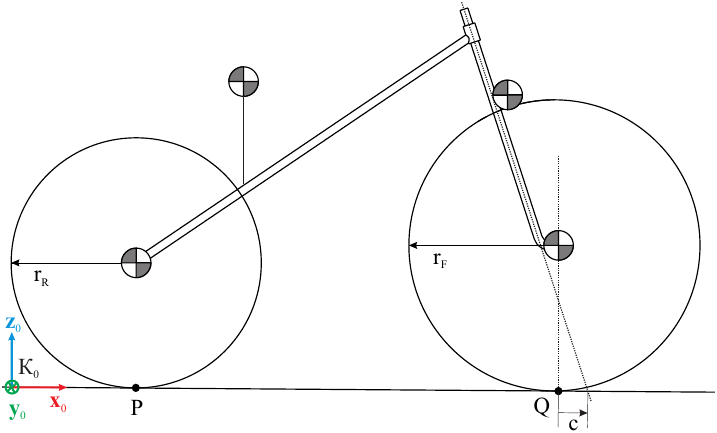}
	\caption{Whipple bicycle shown in the reference configuration where it stands upright without the handlebar being turned. The position of the Center of Mass (COM) of the rear wheel, the rear body, the handlebar, and the front wheel are marked, as well as the two ground contact points \Pp\ and \Qp, the two wheel radii \rR\ and \rF, and the trail $c$ of the bicycle. The global frame, denoted as \Gframe-frame, is shown.}
	\label{fig:BM_characteristics}
\end{figure}

\subsection{Characteristics}
\label{Characteristics}
The Whipple bicycle, shown in Figure \ref{fig:BM_characteristics}, consists of four rigid bodies, namely the Rear wheel (R), the Front wheel (F), the rear Body (B) being the bicycle frame and the rider, and the Handlebar (H) that includes the fork. The bodies are interconnected by three frictionless revolute joints, one at the bicycle head tube and two at the wheel hubs. The Whipple bicycle has a tilted steering axis and a fork offset. Due to the fork offset of the bicycle, the front wheel hub is not located on the steering axis, but at a constant distance to it. The Whipple bicycle has a positive trail $c$. The rider body is assumed to be rigidly connected to the frame of the bicycle. The wheels are modelled using geometrically ideal thin discs, each having one contact point with the flat level ground. The contact points of the wheels with the ground are called \Pp\ and \Qp. The wheels are slip-free, both in the direction of travel and to the side. The geometrical and mechanical parameters of the bicycle model are taken from \cite{Meijaard2007}. The bicycle model is subjected to gravitational acceleration. The bicycle model is self-stable when travelling at a forward velocity between \SI{4.3}{ms^{-1}} and \SI{6}{ms^{-1}} \cite{Meijaard2007}, as initial perturbations of the bicycle model subside over time if moving in this range for the forward velocity.

\subsection{Minimal Coordinates}
\label{Minimal Coordinates}
\begin{figure}[h]
	\centering
	\includegraphics[scale=1]{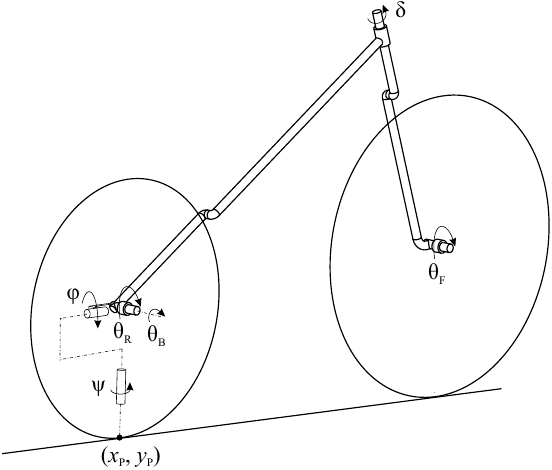}
	\caption{Whipple bicycle drawn with its minimal coordinates on positional base and the pitch angle $\theta_\mathrm{B}$. The upright cylinder marked with $\Psi$ represents the yaw angle, the mounting attached to the rear dropout is used to illustrate the roll angle $\varphi$ that is independent of the pitch angle $\theta_\mathrm{B}$ of the rear body. Note that the pitch angle $\theta_\mathrm{B}$ is not a minimal coordinate of the bicycle, but is needed later for the coordinates mappings.}
	\label{fig:BM_minimalCoordinates}
\end{figure}

To use RL, a state must be defined which describes the environment in such a way that the agent can select a suitable action \cite{Zai2020}. In the present work, the state contains the minimal coordinates of the bicycle model. For the degree of freedom on positional base, $f_\mathrm{L} = 6n - b_\mathrm{L} = 24 - 17 = 7$ follows, where $n$ is the number of rigid bodies of the system and $b_\mathrm{L}$ the number of geometric constraints in the system \cite{Meijaard2007}. The number of geometric constraints is the sum of the number of constraints resulting from the revolute joints plus two, as the wheels cannot penetrate the ground. In Figure \ref{fig:BM_minimalCoordinates} the minimal coordinates on positional base of the system are illustrated. Since the bicycle has non-holonomic constraints, the degree of freedom on velocity base is reduced to $f=f_\mathrm{L} - b^* = 7-4=3$, where $b^*$ is the number of non-holonomic constraints \cite{Meijaard2007}. Per wheel, two non-holonomic constraints are added since neither side-slipping nor free spinning of the wheels is possible. In Table \ref{tab:BM_minimalCoordinates} the minimal coordinates on positional and velocity base are given. The vector $\qv \in \mathbb{R}^{10}$ of the minimal coordinates reads
\begin{equation}
	\qv = \left[
	\begin{array}{cccccccccc}
		x_\Pp & y_\Pp & \Psi & \varphi & \delta & \theta_\mathrm{R} & \theta_\mathrm{F} & \dot{\varphi} & \dot{\delta} & \dot{\theta}_\mathrm{F}
	\end{array}
	\right]\tp \eqdot
	\label{eq:qv}
\end{equation}

\begin{table}
	\tbl{Minimal coordinates of the Whipple bicycle that are part of the state describing the environment. The \TIframe-frame used is defined in Appendix \ref{Coordinates Mappings for the Whipple Bicycle}.}{
		\begin{tabular}{c|p{9cm}}
			\toprule
			\textbf{notation} & \textbf{description} \\
			\midrule
			$x_\Pp$ & x-coordinate of the global position of the rear wheel contact point \\
			$y_\Pp$ & y-coordinate of the global position of the rear wheel contact point \\
			$\Psi$ & yaw angle between the x-axes of the \Gframe-frame and the \TIframe-frame around the z-axis of the \Gframe-frame \\
			$\varphi$ & roll angle around the x-axis of the \TIframe-frame, where $\varphi=0$ means that the z-axes of the \TIframe-frame and the \Gframe-frame are parallel \\
			$\delta$ & steering angle \\
			$\theta_\mathrm{R}$ & rear wheel rotation angle \\
			$\theta_\mathrm{F}$ & front wheel rotation angle \\
			$\dot{\varphi}$ & roll angular velocity \\
			$\dot{\delta}$ & steering angular velocity \\
			$\dot{\theta}_\mathrm{F}$ & front wheel rotation angular velocity \\
			\bottomrule
	\end{tabular}}
	\label{tab:BM_minimalCoordinates}
\end{table}

\subsection{Coordinates Mappings}
\label{Coordinates Mappings}
For the multibody model, a redundant formulation is used, representing the configuration of each rigid body with a translation and rotation. The redundant formulation is used to make the RL approach expandable in the future, that is, to use flexible bodies instead of rigid bodies for the bicycle model or to use an uneven pavement instead of a flat level ground on which the bicycle model drives. By using a redundant formulation, a mapping of the redundant coordinates to the minimal coordinates and vice versa must be established before setting up the RL framework. In Table \ref{tab:BM_coordinatesMappings} the two mappings are described, as well as why the two mappings are required.

\begin{table}[h]
	\tbl{Description of the coordinates (coord.) mappings that must be established for using the multibody model of the bicycle in the RL framework.}{
		\begin{tabular}{p{6cm}|p{6cm}}
			\toprule
			\hfil \textbf{redundant coord. $\rightarrow$ minimal coord.} & \hfil \textbf{minimal coord. $\rightarrow$ redundant coord.} \\
			\midrule
			compute the minimal coord. (on positional and velocity base) from the current configuration of the bicycle since the minimal coord. are part of the state vector describing the environment; without this transformation the state of the environment could not be computed & compute the redundant coord. (on positional and velocity base) when the environment is reset since the environment is reset by randomly initialising the minimal coord. of the bicycle model; without this transformation the simulation could not restart after the environment is reset \\ 
			\bottomrule
	\end{tabular}}
	\label{tab:BM_coordinatesMappings}
\end{table}

In Appendix \ref{Coordinates Mappings for the Whipple Bicycle}, the mapping of the redundant to the minimal coordinates with sensors that are defined in the multibody simulation code and the mapping of the minimal to the redundant coordinates are shown. The mappings include both the positional and velocity base.

\subsection{Model of the Steering Drive}
\label{Model of the Steering Drive}
The agent does path following and stabilisation of the bicycle model by predicting the steering angle $\delta_\mathrm{set}$ of the bicycle model. The prediction for the steering angle is converted into a steering torque $\tau$ by a PD controller. The steering torque $\tau$ is imprinted between the rear body and the fork of the bicycle model and acts around the steering axis, modelling a drive sitting in the bicycle head. The steering torque $\tau$ reads
\begin{equation}
	\tau = P(\delta - \delta_\mathrm{set}) + D(\dot{\delta}-\dot{\delta}_\mathrm{set}) \eqcomma
\end{equation}
where $\delta$ is the steering angle of the bicycle model and $\dot{\delta}$ is the steering angular velocity. For the set value of the steering angular velocity $\dot{\delta}_\mathrm{set}=0$ applies, as the agent sets the steering angle without specifying the steering angular velocity $\dot{\delta}$. The two parameters of the PD controller are set $P=\SI{9}{Nm}$ and $D=\SI{1.6}{Nms}$. In Appendix \ref{Threshold Value for the Steering Angle of the Whipple Bicycle and Step Response of the Steering Angle} it is described how the values for $P$ and $D$ are found.

\section{Reinforcement Learning Framework}
\label{Reinforcement Learning Framework}
\begin{figure}[h]
	\centering
	\includegraphics[scale=1]{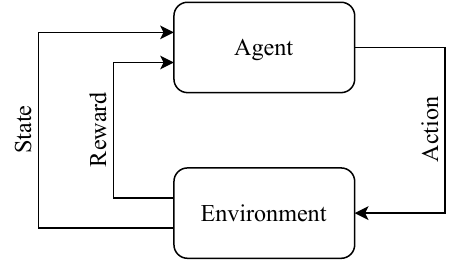}
	\caption{Scheme of the RL framework. An agent can choose an action based on the state of a dynamic environment. A numerical reward signal is passed pack to the agent.}
	\label{fig:RL_RLScheme}
\end{figure}

In an RL framework, the agent can choose an action based on the state of the environment to solve a defined control task. The action chosen by the agent changes the dynamic environment. The new state and a reward are passed back to the agent, closing the agent-environment loop, see Figure \ref{fig:RL_RLScheme}. Based on the new state, the agent again selects an action. The reward is a numerical signal. The agent tries to maximise the cumulative reward and should thus learn to solve the control task. Learning means that the data obtained when passing through the agent-environment loop is used to adapt the behaviour of the agent. The method used to adapt the behaviour is defined by the specific RL algorithm used. In the present work, mainly the off-policy Soft Actor Critic (SAC) algorithm is used, first introduced in the work \cite{Haarnoja2018}. The SAC algorithm is off-policy, such that it uses a replay buffer, where the data obtained when running through the RL framework is saved. After every single run through the RL framework, a batch sampled from the replay buffer is used to adapt the behaviour of the agent \cite{Haarnoja2018}. If the agent is implemented using neural networks, the concept of RL is extended to Deep Reinforcement Learning (DRL). Adapting the behaviour of the agent in the context of DRL means that the weights and biases in the neural networks are adapted \cite{Zai2020}. In the present work, the terms RL and DRL are used synonymously, as all agents are implemented using neural networks. If an agent is deployed, it can be used in a similar way to a controller. To deploy an agent when using the SAC algorithm means that at a selected point during the learning process a neural network of the agent is saved. The saved neural network is the approximator for the state value function, predicting an action to be executed depending on the state of the environment \cite{Haarnoja2018}. When deployed, the agent no longer needs the reward signal and, at best, solves the control task.

The framework set up for the present work is Python (version 3.11.5) based. The multibody simulation code \EXU\footnote{\url{https://github.com/jgerstmayr/EXUDYN} (used version: 1.8.0)} is used for the multibody simulation of the bicycle model. \EXU\ is based on C++ and available as Python package \cite{Gerstmayr2023}. \EXU\ provides an interface class that connects the multibody system to OpenAIGym\footnote{\url{https://github.com/Farama-Foundation/Gymnasium} (used version: 0.21.0)} \cite{Brockman2016}. OpenAIGym in turn provides a standardised port to already existing implementations of RL algorithms. The RL algorithms used in the present work are part of \SB\footnote{\url{https://github.com/DLR-RM/stable-baselines3} (used version: 1.7.0)} \cite{Raffin2021}, which provides implementations of seven different RL algorithms.

In the following, four main parts of the RL framework are described, namely the environment, the state, the reward, and the action. At certain points, several possible settings for the RL framework are given. The different settings for the RL framework are compared to each other later in the present work.

\subsection{Environment}
\label{Environment}
\begin{figure}[h]
	\centering
	\resizebox*{5cm}{!}{\includegraphics[scale=1]{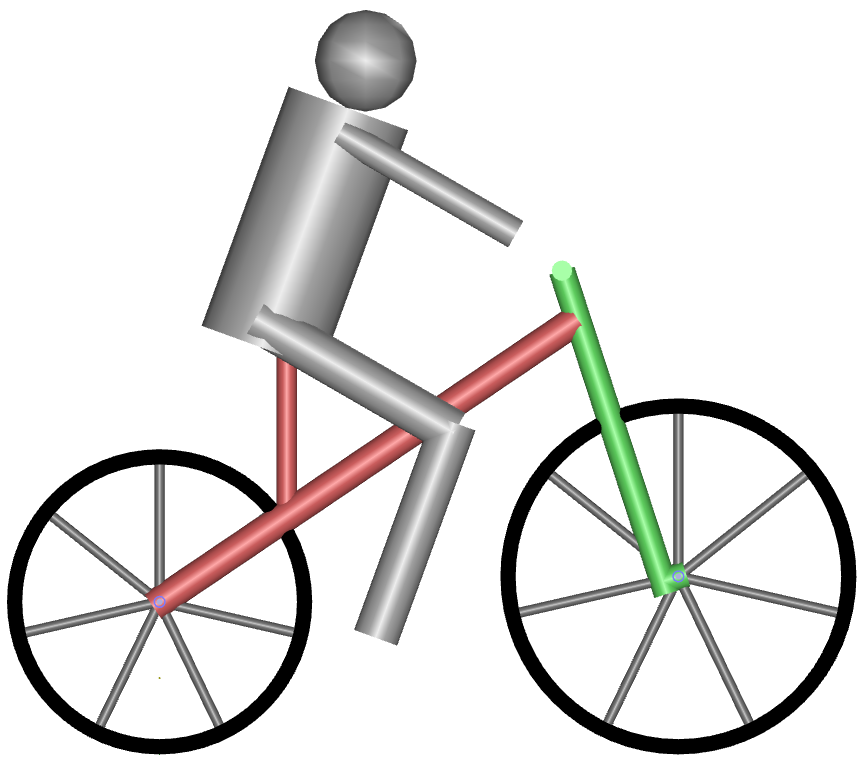}}
	\hspace{5pt}
	\resizebox*{5cm}{!}{\includegraphics[scale=1]{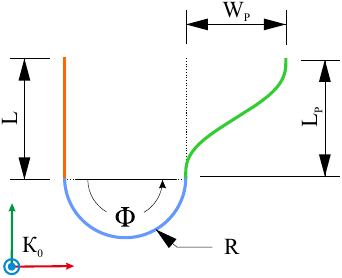}}
	\hspace{5pt}
	\caption{(Left) The visualisation of the virtual bicycle model is shown. (Right) Examples of the three element types with which the paths in the present work are constructed are drawn with their corresponding geometric properties.}
	\label{fig:RL_environment}
\end{figure}

The environment consists of the simulated bicycle model and the path the bicycle should follow. As outlined in Section \ref{Coordinates Mappings}, the multibody model of the bicycle is implemented using a redundant formalism. The differential algebraic equations resulting from the redundant formalism and assembled by \EXU\ are solved using the generalised-$\alpha$ solver implemented in \EXU. The step size of the solver is set \SI{0.005}{s}. The RL algorithm interacts every $h=\SI{0.05}{s}$ with the environment, with $1/h$ giving what is known as controller frequency in control engineering. Thus, $10$ simulation steps are made until a new action chosen by the agent is applied to the multibody system. The update time $h$ is set higher than the step size of the solver so that the dynamics of the environment can be learnt in fewer interactions between the RL algorithm and the environment. The visualisation of the bicycle model is shown in Figure \ref{fig:RL_environment}, left.

\begin{figure}
	\centering
	\makebox[\textwidth][c]{\includegraphics[scale=1]{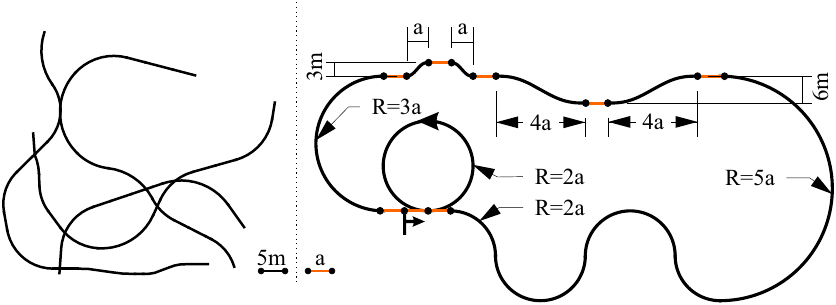}}
	\caption{(Left) Four examples of randomly generated paths which are used in the learning process are shown. Short parts of the paths are drawn each to prevent clustering in the figure. (Right) The benchmark path that is used to evaluate the performance of the agents is shown. For the benchmark path, $a=\SI{5}{m}$ applies.}
	\label{fig:RL_pathTypes}
\end{figure}

The paths the bicycle model should follow are built using three different element types, namely linear elements, polynomial elements, and circular elements, see Figure \ref{fig:RL_environment}, right. The linear element is defined by its length $L$. The circular element is defined using an angle $\Phi$ and a radius $R$. The polynomial element is defined by its length $L_\mathrm{P}$ and width $W_\mathrm{P}$. The polynomial elements are all of degree five. In the present work, two families of paths are used, see Figure \ref{fig:RL_pathTypes}. For the learning of the agents, paths are initialised randomly in a wide range using the reset procedure, so that the agents not only learn to follow a specific path, but to follow paths as generally as possible. The reset procedure is described later in detail. For the performance evaluation of the deployed agents, additionally, a benchmark path is introduced, including a full circle, a slalom, curves, two lane changes, and straight elements.

\subsection{State}
\label{State}
The state in an RL framework includes important geometric and kinematic quantities of the environment which are essential for the agent. The state in the present work contains the minimal coordinates of the bicycle model and preview information describing the path the bicycle should follow. Both parts are described below.

In the preparatory work associated with the bicycle model, its minimal coordinates $\qv$ are defined, see Section \ref{Minimal Coordinates} and Equation \eqref{eq:qv}. One part of the state vector $\sv$ is the vector $\qv$. An alternative redundant formulation $\qv' \in \mathbb{R}^{11}$ of the minimal coordinates is introduced, that can be used instead of the vector $\qv$. In the alternative formulation, the yaw angle $\Psi$ is represented using a unity vector with the two components
\begin{equation}
	x_\mathrm{\Psi} = \cos{\Psi} \qquad \text{and} \qquad y_\mathrm{\Psi} = \sin{\Psi}\eqdot
\end{equation}
Consequently, the alternative redundant formulation for $\qv$ reads
\begin{equation}
	\qv' = \left[
	\begin{array}{ccccccccccc}
		x_\Pp & y_\Pp & x_\mathrm{\Psi} & y_\mathrm{\Psi} & \varphi & \delta & \theta_\mathrm{R} & \theta_\mathrm{F} & \dot{\varphi} & \dot{\delta} & \dot{\theta}_\mathrm{F}
	\end{array}
	\right]\tp \eqdot
	\label{eq:qv_alt}
\end{equation}

\begin{figure}
	\centering
	\includegraphics[scale=1]{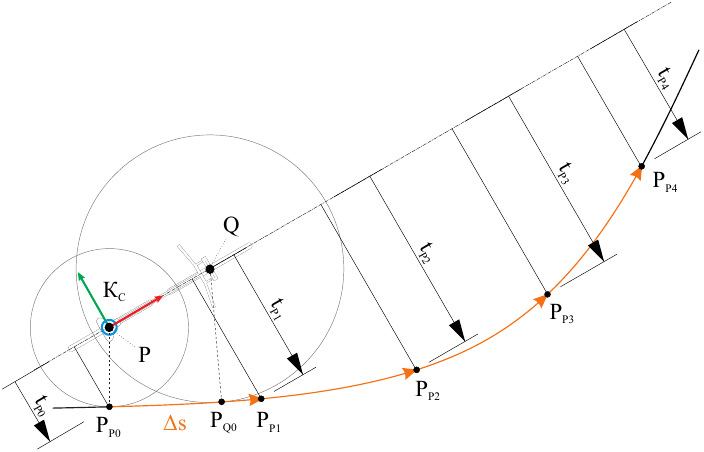}
	\caption{Scheme of the procedure to get preview information of the path. The preview points required for the preview information are obtained by moving along the path by the preview distance $\Delta s$, starting with the point $\mathrm{P_{P0}}$. The point $\mathrm{P_{P0}}$ is the point on the path nearest to the rear wheel contact point \Pp. The point $\mathrm{P_{Q0}}$ is the point on the path nearest to the front wheel contact point \Qp.}
	\label{fig:RL_pathPreview}
\end{figure}

The preview information of the path is given to the agent using the vector $\tv_\Pp \in \mathbb{R}^{(n_\mathrm{prev}+1)}$, where $n_\mathrm{prev}=4$ is the number of preview points used to compute the vector $\tv_\Pp$. The vector $\tv_\Pp$ represents the course of the path ahead of the bicycle model given in the \Cframe-frame. The \Cframe-frame is defined in Equation \eqref{eq:T0T1}. To obtain the vector $\tv_\Pp$, it is started with the point $\mathrm{P_{P0}}$ illustrated in Figure \ref{fig:RL_pathPreview}. The vector pointing from the rear wheel contact point to $\mathrm{P_{P0}}$ is projected on the y-axis of the \Cframe-frame. The length of the projection is the first entry of the vector $\tv_\Pp$, denoted as $t_\mathrm{P0}$. The point $\mathrm{P_{P0}}$ is moved along the path by the preview distance $\Delta s$. This gives the first preview point, called $\mathrm{P_{P1}}$. Again, the vector that is pointing from \Pp\ to $\mathrm{P_{P1}}$ is projected on the y-axis of the \Cframe-frame, which gives the second entry $t_\mathrm{P1}$ of the vector $\tv_\Pp$. Repeating this procedure, the vector $\tv_\Pp$ finally reads
\begin{equation}
	\tv_\Pp = \left[
	\begin{array}{ccccc}
		t_\mathrm{P0} & t_\mathrm{P1} & t_\mathrm{P2} & t_\mathrm{P3} & t_\mathrm{P4}
	\end{array}
	\right]\tp \eqdot
\end{equation}
Taken together, the state vector $\sv$ follows
\begin{equation}
	\sv = \vp{\qv}{\tv_\Pp} \qquad \text{or} \qquad \sv=\vp{\qv'}{\tv_\Pp} \eqdot
\end{equation}
The two formulations for the state vector $\sv$ are compared to each other in the section on the results, as are two ways to compute the preview distance $\Delta s$.

\subsection{Reward}
\label{Reward}
The agent should solve the task of doing path following and stabilisation of the bicycle model. To represent the two components of the task, the reward $\rho \in [0, 1]$ is the linear combination of the two parts $\rho_\mathrm{y}$ and $\rho_\mathrm{\varphi}$.
\begin{equation}
	\rho = \chi_1 \cdot \rho_\mathrm{y} + \chi_2 \cdot \rho_\mathrm{\varphi}
	\label{eq:rho}
\end{equation}
For the two reward weights $\chi_1+\chi_2=1$ applies.

The first part $\rho_\mathrm{y} \in [0, 1]$ rewards the agent when the bicycle is close to the path. A distance referred to as $t_\mathrm{0}$ is used. The distance $t_\mathrm{0}$ can either be the Euclidean distance $d(\Pp, \mathrm{P_{P0}})$ between the rear wheel contact point \Pp\ and $\mathrm{P_{P0}}$ or the Euclidean distance $d(\Qp, \mathrm{P_{Q0}})$ between the front wheel contact point \Qp\ and $\mathrm{P_{Q0}}$. The points $\mathrm{P_{P0}}$ and $\mathrm{P_{Q0}}$ are shown in Figure \ref{fig:RL_pathPreview}. A relationship must be introduced to compute the part $\rho_\mathrm{y}$ with the distance $t_\mathrm{0}$. Either a linear relationship reading
\begin{equation}
	\rho_\mathrm{y} = \rho_\mathrm{y}^\mathrm{lin} = \frac{\eta_\mathrm{y}-t_\mathrm{0}}{\eta_\mathrm{y}}
	\label{eq:rhoYLinear}
\end{equation}
or a relationship forming the normalised Gaussian bell curve, reading
\begin{equation}
	\rho_\mathrm{y} = \rho_\mathrm{y}^\mathrm{exp} = \exp{\left( -0.5 t_\mathrm{0}^2 \right)}
	\label{eq:rhoYGauss}
\end{equation}
can be used. The distance threshold $\eta_\mathrm{y}$ is set to \SI{3.5}{m}. As explained later, the environment is reset if the bicycle model drives further away than $\eta_\mathrm{y}$ from the path. Consequently, when the distance between the bicycle model and the path becomes maximum, the part $\rho_\mathrm{y}$ of the reward equals $0$.

The part $\rho_\mathrm{\varphi} \in [0, 1]$ is computed using the roll angle $\varphi$.
\begin{equation}
	\rho_\mathrm{\varphi}= \frac{\eta_\mathrm{\varphi}-|\varphi|}{\eta_\mathrm{\varphi}}
	\label{eq:rhoPhi}
\end{equation}
By computing the part $\rho_\mathrm{\varphi}$ this way, the agent is penalised if the bicycle leans. The roll angle threshold $\eta_\mathrm{\varphi}$ is set to \ang{45}. If the magnitude of the roll angle $\varphi$ exceeds $\eta_\mathrm{\varphi}$, the environment is reset. Consequently, when the magnitude of the roll angle $\varphi$ becomes maximum, the part  $\rho_\mathrm{\varphi}$ of the reward equals $0$. Although this reward part is conflicting with training the agent to follow paths including curves, since with a curve the roll angle $\varphi$ must be non-zero, it will be investigated later whether and to what extent this interferes with the learning progress. Note that this reward part can be ignored by setting the reward weights $\chi_1=1$ and $\chi_2=0$.

The two described options to obtain the distance $t_\mathrm{0}$ will be compared to each other later, as well as using Equation \eqref{eq:rhoYLinear} or Equation \eqref{eq:rhoYGauss} to compute the part $\rho_\mathrm{y}$. Furthermore, the influence of the reward weights $\chi_1$ and $\chi_2$ is analysed, i.e. how considering the part $\rho_\mathrm{\varphi}$ in the reward changes the learning process of the agents.

\subsection{Action}
\label{Action}
The agent does path following and stabilisation of the bicycle model exclusively by taking steering actions. The agent predicts a steering angle $\delta_\mathrm{set}\in[-\ang{70}, \ang{70}]$ that is applied to the bicycle model as described in Section \ref{Model of the Steering Drive}. The interval given for $\delta_\mathrm{set}$ is substantiated in Appendix \ref{Threshold Value for the Steering Angle of the Whipple Bicycle and Step Response of the Steering Angle}.

\section{Learning Process with Curriculum Learning}
\label{Learning Process with Curriculum Learning}
The learning process runs through the RL framework to optimise the weights of the neural networks the agent consists of. The learning process thus is the process that generates data with the multibody simulation environment and uses this data to optimise the function approximators of the agent. The learning processes in the present work consist of $\steps{tot}=4\cdot 10^6$ learning steps, since, as shown later, the validation results no longer change significantly at this level of training. One learning step is one run-through of the RL framework. This means that one learning process uses data generated in $\steps{tot} \cdot h = 55.5$ hours of simulated bicycle rides, with $h$ being the time between two interactions of the agent, see Section \ref{Environment}. The learning process can be divided into episodes. An episode ends when at least one of the conditions listed below is met:
\begin{itemize}
	\item The episode consists of $\steps{max}=1200$ learning steps so that one episode represents a bicycle ride with a maximum duration of $\steps{max} \cdot h = \SI{60}{s}$.
	\item At least one of the two thresholds $\eta_\mathrm{y}$ and $\eta_\mathrm{\varphi}$ is exceeded, i.e. the environment is reset, and the episode consists of at least $\steps{min}=800$ learning steps. This number of learning steps is determined by the ratio $\steps{min} \cdot 1.5 = \steps{max}$ defined for the present work. The two thresholds $\eta_\mathrm{y}$ and $\eta_\mathrm{\varphi}$ are defined in Section \ref{Reward}.
\end{itemize}
As long as the learning process does not contain \steps{tot} learning steps, a new episode starts when the previous episode ended.

In addition to running through the RL framework, the learning process must reset the environment when certain conditions are met. Furthermore, validations are performed during the learning process to monitor the progress of learning. During the learning process, a training strategy called Curriculum Learning (CL) is applied. The reset procedure, the validation procedure, and the CL are explained now.

\subsection{Resetting the Environment}
\label{Resetting the Environment}
Primarily, the environment of the RL framework must be reset if the agent failed in solving the control task. The reset procedure is therefore called when the roll angle $\varphi$ of the bicycle model exceeds the roll angle threshold $\eta_\mathrm{\varphi}$ or when the distance between the rear wheel and the point  $\mathrm{P_{P0}}$ exceeds the distance threshold $\eta_\mathrm{y}$. In other words, reset is called when the bicycle has fallen over or driven far away from the path. If two consecutive episodes contain the maximum number of learning steps \steps{max}\ and no reset was called between the episodes, the environment is also reset. Resetting the environment at this condition ensures that the agent that already succeeds in solving the control task without exceeding the thresholds $\eta_\mathrm{y}$ and $\eta_\mathrm{\varphi}$ still finds a variety of initial conditions in the remaining learning process. The reset procedure initialises the configuration of the bicycle model as well as the path the bicycle model should follow.

The configuration of the bicycle model is initialised by randomly setting the minimal coordinates of the bicycle model in the intervals given in the upper part of Table \ref{tab:LP_initialisationOfParameters}. The roll angle $\varphi$, the steering angle $\delta$, and their angular velocities are not set to zero when initialising the bicycle model. By initially perturbing the bicycle model, the agent should learn how to stabilise the bicycle by just observing the beginnings of every simulated bicycle ride. The forward velocity of the bicycle model is set randomly between $v_\mathrm{min}$ and $v_\mathrm{max}$. Note that the bicycle cannot accelerate or decelerate in the simulated bicycle rides, as required in the introductory Section \ref{Introduction}. Since the forward velocity is not a minimal coordinate of the bicycle model, it is converted into the front wheel rotation angular velocity $\dot{\theta}_\mathrm{F}$ using
\begin{equation}
	\dot{\theta}_\mathrm{F} = \frac{v}{\rF} \eqcomma
\end{equation}
with \rF\ being the radius of the front wheel. See Appendix \ref{Coordinates Mappings for the Whipple Bicycle} to see how the forward velocity is considered in the coordinates mappings. The parameters not necessary for solving the control task, i.e. the contact point $\Pp=(x_\Pp, y_\Pp)$, the yaw angle $\Psi$, and the wheel rotation angles $\theta_\mathrm{R}$ and $\theta_\mathrm{F}$, are initialised arbitrarily so that the agent learns their irrelevance. Remember that the used multibody simulation of the bicycle model is based on a redundant formulation, which is why the minimal coordinates must be mapped to redundant coordinates after calling reset, as elaborated in Section \ref{Coordinates Mappings}.

The randomly initialised path the bicycle model should follow lies in the plane spanned by the x-axis and the y-axis of the \Gframe-frame. The beginning of the path is placed where the contact point \Pp\ of the rear wheel is initialised. The beginning of the path is rotated such that the angle of the tangent of the path at the point \Pp\ is inside the interval $[\Psi-\ang{5}, \Psi+\ang{5}]$, where $\Psi$ is the yaw angle the bicycle is initialised with. The path is initialised by appending a random sequence of the three element types the path can consist of. The three element types are explained in Section \ref{Environment} and illustrated in Figure \ref{fig:RL_environment}. Not only the sequence of element types, but also the geometric properties of each individual element of the path are initialised randomly in a given interval. The intervals for the geometric properties of the elements are given in the lower part of Table \ref{tab:LP_initialisationOfParameters}. The curvature of the circular elements is selected at random so that with a probability of \SI{50}{\percent} the element is positively or negatively curved. The same is done for the polynomial elements to specify the direction of the transfer of the width $|W_\mathrm{P}|$. It is specified that the initialised path does not start with a circular element so that after resetting, where the bicycle model may be perturbed, the task does not become unachievable. Additionally, it is defined that \SI{40}{\%} of the elements in the initialised path should be circular elements, \SI{40}{\%} should be polynomial elements, and \SI{20}{\%} should be straight elements. This is specified so that the path does not contain long straight sections with which the agent exploits a behaviour being able to follow the straight elements, but not the other two types of elements. The family of paths assembled in the way of the present work cannot be followed by the bicycle model exactly, because the dynamic characteristics of the bicycle impose limitations in its manoeuvrability \cite{Astrom2005}. In particular, step-like changes in the curvature of the path result in a path error between the bicycle model and the path, alike a bicycle travelling on real roads. To start the turn of a bicycle, the handlebar must be turned in the opposite direction, known as counter-steering, so that the bicycle leaves the desired path \cite{Meijaard2007, Astrom2005}. This dynamic property of the bicycle model must be learnt by the agents, trying to follow the path as closely as possible.

\begin{table}[h]
	\tbl{Intervals in which the parameters of the environment are initialised when reset is called, where for each parameter $x$ an interval $a\leq x \leq b$ is given. The table indicates which initialisation intervals are used to apply CL, see also Table \ref{tab:LP_initialisationOfCurriculumParameters}.}{
		\begin{tabular}{c|cc|c}
			\toprule
			\multirow{2}{*}{\textbf{parameter}} & \multicolumn{2}{c|}{\textbf{initialisation interval}} & \multirow{2}{*}{\textbf{CL}} \\
			& $a$ &  $b$ & \\
			\cmidrule{1-4}
			\multicolumn{4}{c}{bicycle model} \\
			\cmidrule{1-4}
			coordinates $x_\Pp$ and $y_\Pp$ of the contact point \Pp  & \SI{-10}{m} & \SI{10}{m} & no \\
			yaw angle $\Psi$ & $\SI[parse-numbers = false]{-\pi}{rad}$ & $\SI[parse-numbers = false]{\pi}{rad}$ & no \\
			roll angle $\varphi$ & \SI{-0.01}{rad} & \SI{0.01}{rad} & no \\
			steering angle $\delta$ & \SI{-0.01}{rad} & \SI{0.01}{rad} & no \\
			wheel rotation angles $\theta_\mathrm{R}$ and $\theta_\mathrm{F}$ & $\SI[parse-numbers = false]{-\pi}{rad}$ & $\SI[parse-numbers = false]{\pi}{rad}$ & no \\
			roll angular velocity $\dot{\varphi}$ & \SI{-0.05}{rad s^{-1}} & \SI{0.05}{rad s^{-1}} & no \\
			steering angular velocity $\dot{\delta}$ & \SI{-0.01}{rad s^{-1}} & \SI{0.01}{rad s^{-1}} & no \\
			front wheel rotation angular velocity $\dot{\theta}_\mathrm{F}$ & \multicolumn{2}{c|}{set with the forward velocity $v$} & - \\
			forward velocity $v$ & $v_\mathrm{min}$ & $v_\mathrm{max}$ & yes \\
			\cmidrule{1-4}
			\multicolumn{4}{c}{path} \\
			\cmidrule{1-4}
			length of linear elements $L$ & \SI{5}{m} & \SI{15}{m} & no \\
			angle of circular elements $\Phi$ & $\SI[parse-numbers = false]{\pi/4}{rad}$ & $\SI[parse-numbers = false]{\pi/2}{rad}$ & no \\
			radius of circular elements $R$ & $R_\mathrm{min}$ & $R_\mathrm{max}$ & yes \\
			length of polynomial elements $L_\mathrm{P}$ & $L_\mathrm{P, min}$ & $L_\mathrm{P, max}$ & yes \\
			width of polynomial segments $W_\mathrm{P}$ & $W_\mathrm{P, min}$ & $W_\mathrm{P, max}$ & yes \\
			\bottomrule
	\end{tabular}}
	\label{tab:LP_initialisationOfParameters}
\end{table}

\subsection{Validation during the Learning Process}
\label{Validation during the Learning Process}
The validation procedure is to monitor the success of the learning process. In the learning process, a validation is always carried out after an episode with \steps{max}\ learning steps and if the last validation was $4000$ learning steps ago. The first condition is set since it is assumed that an agent that manages to complete the maximum number of steps in an episode already masters the task to some extent. The latter condition is set to save computing time, as a well-trained agent is not validated after each episode. A validation consists of $10$ simulated bicycle rides of a duration of \SI{30}{s}. To do path following and stabilisation of the bicycle model in these simulated bicycle rides, the agent in its current state of training is used as a kind of controller. Before each of the simulated bicycle rides in the validation, the environment is reset using the procedure and initialisation intervals described in Section \ref{Resetting the Environment}. Each simulated bicycle ride is scored using a normalised error $e_i\in[0, 1]$.
\begin{equation}
	e_i=
	\begin{cases}
		1 & \text{if $\eta_\mathrm{y}$ or $\eta_\mathrm{\varphi}$ is exceeded}\\
		\frac{\bar{t}_\mathrm{0, \SI{80}{\%}}}{\eta_\mathrm{y}} & \text{else}
	\end{cases}
\end{equation}
If the bicycle model has fallen over or driven far away from the path, which means that the distance threshold $\eta_\mathrm{y}$ or the roll angle threshold $\eta_\mathrm{\varphi}$ is exceeded, $e_i=1$ is set. Otherwise, the error $e_i$ is computed using the mean of $t_\mathrm{0}$ measured over the last \SI{80}{\%} of the duration of the simulated bicycle ride, denoted as $\bar{t}_\mathrm{0, \SI{80}{\%}}$. The first \SI{20}{\%} of the bicycle ride are ignored for this computation, in order to make $e_i$ independent of the perturbations the bicycle model is initialised with. Remark that the distance $t_\mathrm{0}$ is calculated the same as for the reward $\rho$, which is described in Section \ref{Reward}. An error threshold $\eta_\mathrm{e}$ is introduced, reading
\begin{equation}
	\eta_\mathrm{e}=\frac{\SI{0.2}{m}}{\eta_\mathrm{y}} = \frac{\SI{0.2}{m}}{\SI{3.5}{m}} \approx 0.057 \eqdot
\end{equation}
In a validation process, a single simulated bicycle ride is called successful if $e_i\leq \eta_\mathrm{e}$ holds. Since a validation consists of $10$ simulated bicycle rides, a validation results in $10$ errors $e_i$.

\subsection{Curriculum Learning}
\label{Curriculum Learning}
Curriculum Learning (CL) is a training strategy for the field of machine learning that was first formalised in the work \cite{Bengio2009}. When doing CL, the task to be solved becomes increasingly complex as the learning proceeds in order to improve the quality of the learnt policy and the speed of the learning process as compared to learning the complex task directly. The CL is applied in the present work by changing the initialisation interval of some parameters, marked in Table \ref{tab:LP_initialisationOfParameters}, throughout learning.

\begin{table}[b]
	\tbl{Intervals in which the parameters that are changed throughout the learning process are initialised in each part of the four-part curriculum, where for each parameter $x$ an interval $a\leq x \leq b$ is given.}{
		\begin{tabular}{c|cc|cc|cc|cc}
			\toprule
			\multirow{3}{*}{\textbf{parameter}} & \multicolumn{8}{c}{\textbf{initialisation range at curriculum part}} \\
			& \multicolumn{2}{c|}{\textbf{1}} & \multicolumn{2}{c|}{\textbf{2}} & \multicolumn{2}{c|}{\textbf{3}} & \multicolumn{2}{c}{\textbf{4}} \\
			& a & b & a & b & a & b & a & b \\
			\midrule
			forward velocity $v$ in \si{ms^{-1}} & $4$ & $4$ & $2$ & $7$ & $2$ & $7$ & $2$ & $7$ \\
			radius of circular elements $R$ in \si{m} & $14$ & $14$ & $12$ & $14$ & $10$ & $14$ & $8$ & $14$ \\
			length of polynomial elements $L_\mathrm{P}$ in \si{m} & $20$ & $22$ & $18$ & $22$ & $16$ & $22$ & $14$ & $22$ \\
			width of polynomial elements $W_\mathrm{P}$ in \si{m} & $2$ & $2$ & $2$ & $4$ & $2$ & $6$ & $2$ & $8$ \\
			\bottomrule
	\end{tabular}}
	\label{tab:LP_initialisationOfCurriculumParameters}
\end{table}

\begin{figure}
	\centering
	\includegraphics[scale=1]{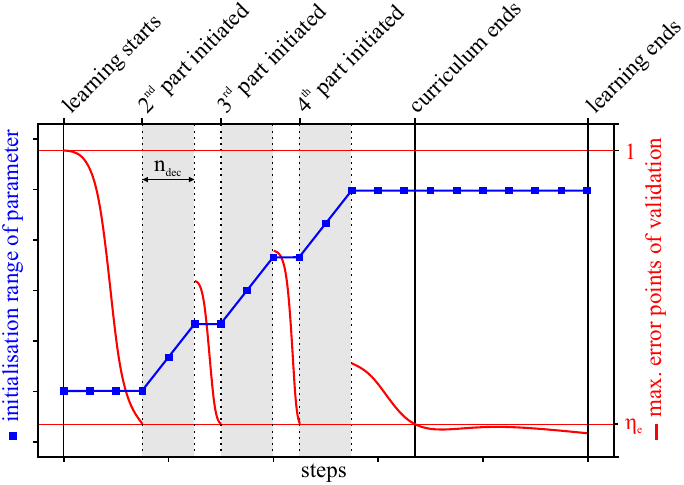}
	\caption{Scheme of a four-part curriculum in which the initialisation range of one parameter is adjusted. When a validation is fully successful, with all errors $e_i \leq \eta_\mathrm{e}$, the next part of the curriculum is initiated. During changing the initialisation range within \steps{dec}\ learning steps, no validation is done. A fully successful validation in the last part of the curriculum represents the end of the curriculum.}
	\label{fig:LP_CLScheme}
\end{figure}

The adaption of the initialisation intervals is done using discrete stages, called parts of the curriculum. The functionality of the curriculum applied in the present work is schematically illustrated in Figure \ref{fig:LP_CLScheme}. A linear relationship over $\steps{dec}=10000$ learning steps is used in the present work to extend the initialisation intervals. In Table \ref{tab:LP_initialisationOfCurriculumParameters} the initialisation intervals that are changed with the parts of the curriculum are given. It can be seen that the curriculum in the present work extends the forward velocity at which the bicycle is initialised from a constant value of \SI{4}{ms^{-1}} to a wide range of speeds. The speed range goes from \SI{2}{ms^{-1}}, a velocity at which the bicycle is in a highly unstable mode, via the self-stable velocity range from \SI{4.3}{ms^{-1}} to \SI{6}{ms^{-1}}, up to \SI{7}{ms^{-1}}, where the bicycle is again mildly unstable \cite{Meijaard2007}. This makes the task more complex in two ways, since the policy must get speed-dependent, see introductory Section \ref{Introduction}, and the slower bicycle is more difficult to stabilise \cite{Meijaard2007}. Furthermore, the curriculum makes the paths to be followed more complex as the learning proceeds by adding sharper turns and steeper transfers to the assembled paths.

\section{Results}
\label{Results}
The section on the results is divided into five parts. Firstly, different settings for the RL framework are listed. Remark that each option for the settings is explained in more detail in Section \ref{Reinforcement Learning Framework}, where the implementation of the RL framework is described. Secondly, the learning process when using the different settings is analysed. Afterwards, the performance of the trained and deployed agents when they are used to do path following and stabilisation of the bicycle model is evaluated. Then, a simulated bicycle ride along the benchmark path is analysed in more detail. Finally, explanatory methods are applied to explain the predictions a trained agent makes. For statistical significance, five learning processes, referred to as \texttt{runs}, initialized with a different random seed, are done with the investigated settings. Therefore, in this section, numerous selection procedures are performed to start with a general analysis and then analyse a single agent as if it were a controller designed using control engineering techniques. 

\subsection{Investigated Settings for the RL Framework}
\label{Investigated Settings for the RL Framework}
The settings for the RL framework concern the state vector $\sv$, the design of the reward $\rho$, and the used RL algorithm.

In the state vector $\sv$, either the vector $\qv$ for the minimal coordinates of the bicycle model, see Equation \eqref{eq:qv}, or a redundant formulation $\qv'$ with the yaw of the bicycle given as a unity vector, see Equation \eqref{eq:qv_alt}, is used. Furthermore, the preview distance $\Delta s$ used to obtain the preview information is either set constant with $\Delta s = \SI{2}{m}$ or scaled with the forward velocity $v$ of the bicycle, reading
\begin{equation}
	\Delta s = v\;\SI{0.4}{s} \eqdot
\end{equation}

The distance $t_\mathrm{0}$ used to compute the part $\rho_\mathrm{y}$ of the reward is either the distance $d(\Pp, \mathrm{P_{P0}})$ using the rear wheel contact point \Pp, or the distance $d(\Qp, \mathrm{P_{Q0}})$ using the front wheel contact point \Qp. With this distance, either a linear relationship $\rho_\mathrm{y}^\mathrm{lin}$, see Equation \eqref{eq:rhoYLinear}, or an exponential relationship $\rho_\mathrm{y}^\mathrm{exp}$, see Equation \eqref{eq:rhoYGauss}, can be used to compute the part $\rho_\mathrm{y}$ of the reward. The reward weights $\chi_1$ and $\chi_2$ are adjustable, see Equation \eqref{eq:rho}. In the present work, either $\chi_1=1$ and $\chi_2=0$ or $\chi_1=\chi_2=0.5$ is used. The latter setting is used to investigate how the inconsistency of the two reward parts $\rho_\mathrm{y}$ and $\rho_\mathrm{\varphi}$, which is described in Section \ref{Reward}, disrupts the learning progress.

While any RL algorithm can be used in the introduced RL framework, as long as a continuous action space is supported by the algorithm, the SAC algorithm \cite{Haarnoja2018} performed significantly better in initial tests than the A2C algorithm introduced in the work \cite{Mnih2016} and the PPO algorithm introduced in the work \cite{Schulman2017}. The results shown were therefore achieved exclusively with the SAC algorithm. The implementation of SAC in \SB\ with the hyperparameters given in \cite{Haarnoja2018} is used.

In Table \ref{tab:R_investigatedSettings}, the investigated settings are listed. Proposed settings are defined, to which each alternative combination of settings differs in one single parameter to better understand their influence on learning and the trained agent.

\begin{table}[h]
	\tbl{Variation of the RL framework. The single parameter that distinguishes the alt(1-5) settings from the proposed (prop.) settings is highlighted.}{
	\begin{tabular}{c|cccccc}
		\toprule
		\textbf{settings} & \textbf{$\qv \vee \qv'$} & \textbf{$\Delta s$} & \textbf{$t_\mathrm{0}$} & \textbf{$\rho_\mathrm{y}$} & \textbf{$\chi_1$} & \textbf{algorithm} \\
		\midrule
		prop. & $\qv'$ & $v\;\SI{0.4}{s}$ & $d(\Pp, \mathrm{P_{P0}})$ & $\rho_\mathrm{y}^\mathrm{lin}$ & $1$ & SAC \\
		alt(1) & \cellcolor{gray!15}$\qv$ & $v\;\SI{0.4}{s}$ & $d(\Pp, \mathrm{P_{P0}})$ & $\rho_\mathrm{y}^\mathrm{lin}$ & $1$ & SAC \\
		alt(2) & $\qv'$ & \cellcolor{gray!15}$\SI{2}{m}$ & $d(\Pp, \mathrm{P_{P0}})$ & $\rho_\mathrm{y}^\mathrm{lin}$ & $1$ & SAC \\
		alt(3) & $\qv'$ & $v\;\SI{0.4}{s}$ & \cellcolor{gray!15}$d(\Qp, \mathrm{P_{Q0}})$ & $\rho_\mathrm{y}^\mathrm{lin}$ & $1$ & SAC \\
		alt(4) & $\qv'$ & $v\;\SI{0.4}{s}$ & $d(\Pp, \mathrm{P_{P0}})$ & \cellcolor{gray!15}$\rho_\mathrm{y}^\mathrm{exp}$ & $1$ & SAC \\
		alt(5) & $\qv'$ & $v\;\SI{0.4}{s}$ & $d(\Pp, \mathrm{P_{P0}})$ & $\rho_\mathrm{y}^\mathrm{lin}$ & \cellcolor{gray!15}$0.5$ & SAC \\
		\bottomrule
	\end{tabular}}
	\label{tab:R_investigatedSettings}
\end{table}

\subsection{Learning Process}
\label{Learning Process}
To analyse the learning process with the different settings for the RL framework, the required learning steps, the progression of the validation error over the learning steps, and the learning steps required in the parts of the curriculum are considered.

\begin{figure}
	\centering
	\includegraphics[scale=1]{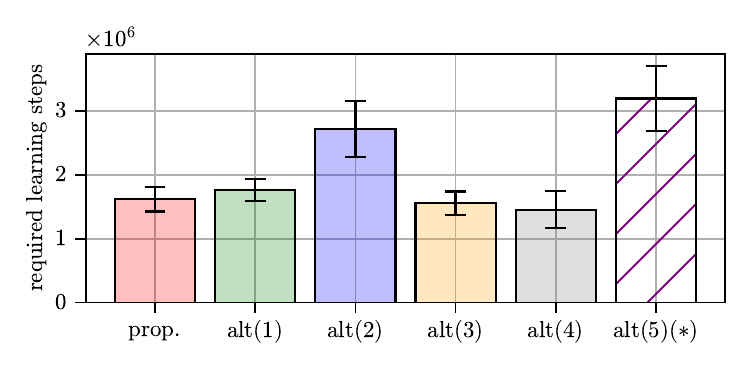}
	\caption{Required learning steps shown with standard deviation when different settings for the RL framework are used. ($\ast$ Not all of the \texttt{runs} done with these settings have reached the end of the curriculum.)}
	\label{fig:R_requiredLearningSteps}
\end{figure}

The number of learning steps required when using specific settings for the RL framework is the mean number of learning steps the \texttt{runs} with these settings take to end the curriculum. As can be seen in Figure \ref{fig:R_requiredLearningSteps}, the alt(2) settings, where the preview distance $\Delta s$ is not scaled with the forward velocity $v$, and the alt(5) settings, where the reward weights $\chi_1$ and $\chi_2$ are set equally, require more than $2.5\cdot 10^6$ learning steps. The proposed and alt(1, 3-4) settings require less than $2\cdot 10^6$ learning steps. With regard to the alt(1) settings, this implies that the yaw representation does not influence the required learning steps. Note that in initial tests the yaw angle $\Psi$ was not initialised randomly, but at $\Psi=0$, with which the curriculum could not be finished regardless of the used yaw representation.

\begin{figure}
	\centering
	\includegraphics[scale=1]{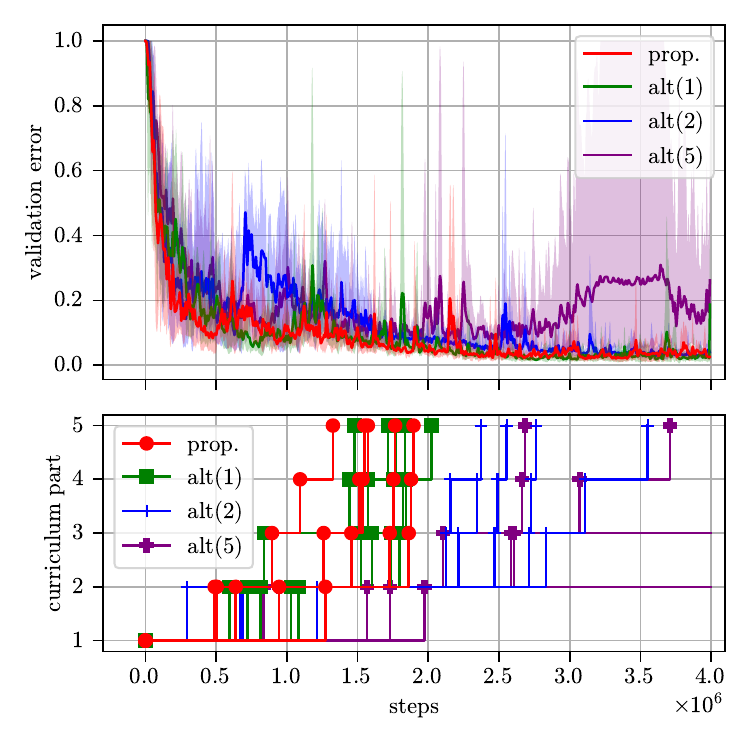}
	\caption{(Top) Validation error $e$ with the proposed and alt(1-2, 5) settings plotted over the learning steps. The mean validation error of the five \texttt{runs} done with each of the settings is drawn, as well as the area between the minimal and maximal validation error. A Simple Moving Average (SMA) filter with $k=20$ is applied to the data. (Bottom) Parts of the curriculum in which the individual \texttt{runs} with the different settings are, plotted over the learning steps.}
	\label{fig:R_learningProcess}
\end{figure}

As described in Section \ref{Validation during the Learning Process}, each validation during the learning process results in $10$ errors $e_i$. To obtain the learning progress, the validation error $e$, being the mean value of the errors $e_i$, is computed over the learning steps for each \texttt{run}. Thus, five errors $e$ can be computed over the learning steps for each setting. In Figure \ref{fig:R_learningProcess} the validation error $e$ is plotted for the proposed and the alt(1-2, 5) settings over the learning steps, as well as the parts of the curriculum in which the \texttt{runs} with these settings are. No significant difference is observed between the proposed and the alt(1) settings, where the yaw of the bicycle is given as an angle value. In the first part of the curriculum, the learning process with the proposed and the alt(2) settings with a non velocity-scaled preview distance $\Delta s$ does not differ significantly, since the velocity scaling of $\Delta s$ has no influence in the first part of the curriculum, with the forward velocity of the bicycle being constant, see Table \ref{tab:LP_initialisationOfCurriculumParameters}. In the second part of the curriculum, the interval for the forward velocity $v$ is extended and the learning processes with the alt(2) settings with a constant preview distance $\Delta s$ take an average of \SI{264}{\%} of the learning steps taken when using the proposed settings. Furthermore, the mean validation error rises above $0.4$ with the alt(2) settings, while the validation error does not exceed $0.26$ with the proposed settings. The learning with the alt(2) settings does not recover from the delay in the second part of the curriculum until the end of the learning process, necessitating the higher number of required learning steps already shown in Figure \ref{fig:R_requiredLearningSteps}. Two learning processes finish the curriculum in \steps{tot} learning steps if the alt(5) settings with equally weighted reward parts are used. The other \texttt{runs} stay in the second and third part of the curriculum until \steps{tot} learning steps are reached. The validation error starts to increase after $2.5 \cdot 10^6$ learning steps, implying that including the part $\rho_\mathrm{\varphi}$ in the reward disrupts the learning. Learning processes with the alt (3) settings, where the front wheel contact point $\Qp$ is used to obtain the distance $t_\mathrm{0}$, and the alt(4) settings, where the exponential relationship $\rho_\mathrm{y}^\mathrm{exp}$ is used, show no significant difference from learning processes with the proposed or the alt(1) settings.

\subsection{Performance of the Deployed Agents}
\label{Performance of the Deployed Agents}
In this section, the performance of the deployed agents is shown, as well as the influence the settings for the RL framework have on the performance of the deployed agents. Remember that a deployed agent takes the place of the controller, applying steering actions based on the current state of the environment to do path following and stabilisation of the bicycle model. Firstly, it is described at which point in a \texttt{run} the agent is deployed and how the performance of the agents is measured. Afterwards, the agents are tested with the family of paths used during learning and along the benchmark path that is described in Section \ref{Environment}.

\subsubsection{Agent Selection and Performance Measure}
\label{Agent Selection and Performance Measure}
The parameters of the agent, i.e. the neural network that predicts the actions, are saved periodically during training. For the evaluation of performance, the last agent from each learning process is chosen which passed all tests of a validation and has finished the curriculum. Consequently, five agents are deployed and evaluated each with the proposed and alt(1-4) settings and two agents are deployed and evaluated with the alt(5) settings. During the simulated bicycle rides using the deployed agents, the distance $t_\mathrm{0}=d(\Pp, \mathrm{P_{P0}})$ is tracked, that is the smallest distance between the rear wheel contact point \Pp\ and the path. The mean number of the measured distance $t_\mathrm{0}$ during the bicycle ride when using the agent $k$ at a forward velocity of $v$ is denoted as $\bar{t}_\mathrm{0, k}(v)$. The maximum of the measured distance $t_\mathrm{0}$ is denoted as $\hat{t}_\mathrm{0, k}(v)$. Both values are used to quantify the performance of the agent $k$. Assuming that the agents $k \in \mathbb{N} \cap [1, N]$ are deployed from the $N$ \texttt{runs} with the proposed settings, the distance $T(v)$ reading
\begin{equation}
	T(v) = \frac{1}{N}\sum_{k=1}^{N} {\bar{t}_\mathrm{0, k}(v)}
\end{equation}
gives the performance of the agents trained with the proposed settings at the forward velocity of $v$. In the following, $T(v)$ is computed for all the settings shown in Table \ref{tab:R_investigatedSettings} with $\SI{2}{ms^{-1}} \leq v \leq \SI{7}{ms^{-1}}$ applying for the forward velocity $v$. Note that during the simulated bicycle rides, the same settings apply as during learning. If, for example, an agent is trained with a non velocity-scaled preview distance, it is also evaluated with the preview distance being non velocity-scaled.

\subsubsection{Performance along Randomly Generated Paths}
\label{Performance along Randomly Generated Paths}
\begin{figure}
	\centering
	\makebox[\textwidth][c]{\includegraphics[scale=1]{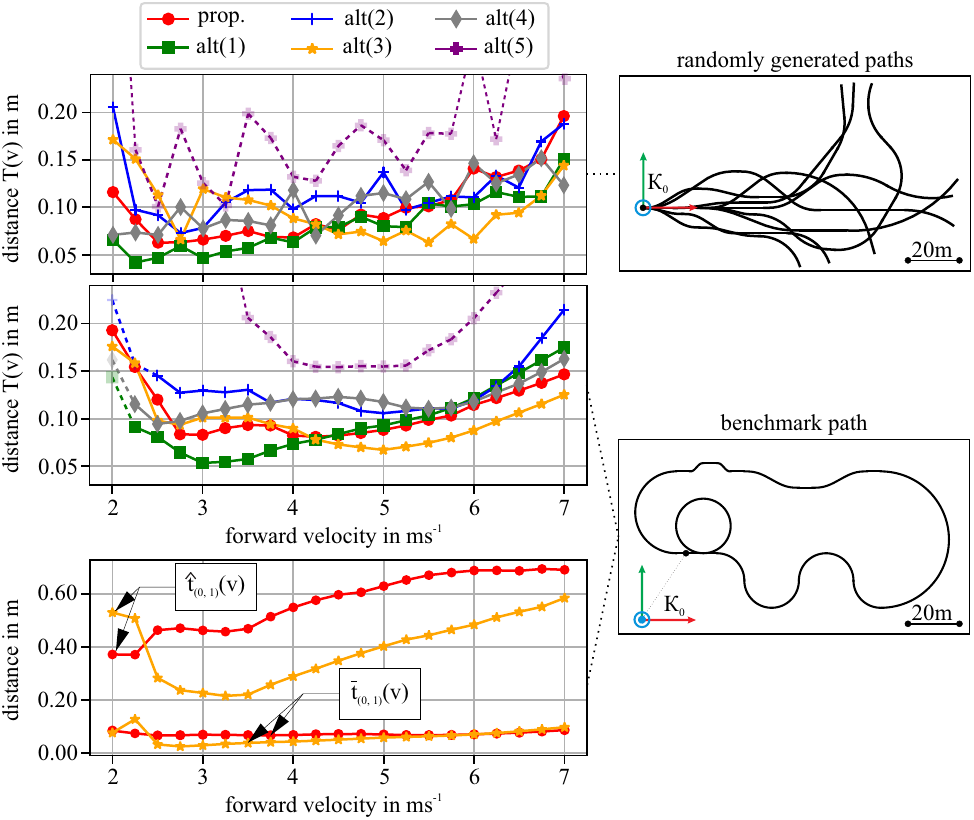}}
	\caption{(Top, Middle) Distance $T(v)$ plotted over the forward velocity $v$ for the investigated settings when two different types of paths are used (schematic representation of the paths shown). (Bottom) Two individual agents are analysed in more detail. The mean $\bar{t}_\mathrm{0, 1}(v)$ and maximal distance $\hat{t}_\mathrm{0, 1}(v)$ are plotted, giving the performance of the two agents over the investigated velocity range along the benchmark path.}
	\label{fig:R_errorOverVelocity}
\end{figure}

The randomly generated paths for evaluation always start in the origin of the \Gframe-frame. The bicycle model is initialized in the reference configuration, see Figure \ref{fig:BM_characteristics}, with the rear wheel contact point \Pp\ on the path to avoid inconsistencies in the performance due to different initialisations. The duration of the simulated bicycle rides is set to \SI{30}{s}. In the upper part of Figure \ref{fig:R_errorOverVelocity}, the distance $T(v)$ is plotted for different settings. If one or more of the five agents that can be deployed from one setting fails at a specific forward velocity $v$, the corresponding marker in the plot is shown translucent and connected by dotted lines. From the outset, this is the case for all markers of the alt(5) settings since only two agents finish the curriculum and are thus deployed. It is seen that all of the agents trained using the proposed and the alt(1-4) settings are able to do path following and stabilisation of the bicycle model along the randomly generated paths over the entire speed range. Taken together, for the evaluation of the proposed and alt(1-4) settings $525$ simulated bicycle rides are made in total ($21$ velocities are tested with the agents deployed from the five \texttt{runs} for each of the five settings) and in all $525$ bicycle rides the bicycle is stabilised and steered along the paths. In other words, none of the agents trained with the proposed and alt(1-4) settings fails in solving the control task for the type of paths the agents are trained with. Note that for each simulated bicycle ride the path is initialized randomly. The distance $T(v)$ for the agents trained with the proposed settings is below \SI{0.2}{m} over the investigated speed range. The same applies for the alt(1, 3-4) settings.

\subsubsection{Performance along the Benchmark Path}
\label{Performance along the Benchmark Path}
The bicycle model is again initialised in the reference configuration, see Figure \ref{fig:BM_characteristics}, with the rear wheel contact point \Pp\ standing at the starting point of the benchmark path. The simulated bicycle rides along the benchmark path do not have a time limit, but the simulations stop when the bicycle returns to the start point. Figure \ref{fig:R_errorOverVelocity} shows that all of the agents trained with the proposed settings manage to do path following and stabilisation of the bicycle model along the benchmark path over the entire velocity range. The same applies to the agents trained with the alt(3) settings, where the reward is computed using an exponential relationship $\rho_\mathrm{y}^\mathrm{exp}$. The distance $T(v)$ when using the proposed settings is below \SI{0.2}{m} over the entire speed range. At \SI{4.25}{ms^{-1}} the distance $T$ equals \SI{0.08}{m}, being the smallest distance $T(v)$ for the proposed settings. It is found that at the lower end of the investigated velocity range, $T$ reads $\SI{0.19}{m}$ and at the upper end $T$ equals $\SI{0.15}{m}$. One agent each deployed using the alt(1-2) settings does not succeed at \SI{2}{ms^{-1}}. One agent using the alt(4) does not succeed at \SI{2}{ms^{-1}} and at \SI{2.25}{ms^{-1}}. This fact will be part of the closing discussion of the present work.

With the proposed and the alt(3) settings one deployed agent each is now analysed in more detail. To select the best agent from the agents deployed when using these two settings, the maximal distance $\hat{t}_\mathrm{0, k}(v)$ is obtained for every agent $k$ deployed from the settings. The value for $\hat{t}_\mathrm{0, k}(v)$ is maximal at a certain forward velocity. The agent with the smallest maximal value for $\hat{t}_\mathrm{0, k}(v)$ is selected. Doing so, for the proposed settings and the alt(3) settings one agent, referred to as \texttt{agent 1} ($k=1$) is chosen each. In the bottom part of Figure \ref{fig:R_errorOverVelocity} the distance values $\hat{t}_\mathrm{0, 1}(v)$ and $\bar{t}_\mathrm{0, 1}(v)$ are plotted over the velocity when these agents are used to do path following and stabilisation of the bicycle model. Remember that the figure shows what can be seen as the performance of the controller depending on the forward velocity of the bicycle, with the mean distance to the benchmark path and the maximal distance being given. Table \ref{tab:R_resultsSingleAgents} shows the maximal values of these two distances when using the two selected agents.

\begin{table}[h]
	\tbl{Maximal values for $\hat{t}_\mathrm{0, 1}(v)$ and $\bar{t}_\mathrm{0, 1}(v)$ when the deployed agents are used to do path following and stabilisation of the bicycle model along the benchmark path. The velocities at which the maximal values occur are given.}{
	\begin{tabular}{c|c|c}
		\toprule
		\textbf{name} & \textbf{value in \SI{}{m}} & \textbf{velocity $v$ in \SI{}{ms^{-1}}} \\
		\cmidrule{1-3}
		\multicolumn{3}{c}{\texttt{agent 1} trained with the proposed settings} \\
		\cmidrule{1-3}
		maximal value for $\hat{t}_\mathrm{0, 1}(v)$ & $0.69$ & $7$ \\
		maximal value for $\bar{t}_\mathrm{0, 1}(v)$ & $0.09$ & $7$ \\
		\cmidrule{1-3}
		\multicolumn{3}{c}{\texttt{agent 1} trained with the alt(3) settings} \\
		\cmidrule{1-3}
		maximal value for $\hat{t}_\mathrm{0, 1}(v)$ & $0.58$ & $7$ \\
		maximal value for $\bar{t}_\mathrm{0, 1}(v)$ & $0.13$ & $2.25$ \\
		\bottomrule
	\end{tabular}}
	\label{tab:R_resultsSingleAgents}
\end{table}

Taken together, an agent is trained with the proposed settings for the RL framework that can stabilise the bicycle model in the velocity range from \SIrange{2}{7}{ms^{-1}} while the rear wheel contact point \Pp\ is never more than \SI{0.69}{m} away from the benchmark path. The largest mean distance to the benchmark path when using this agent as a kind of controller for the bicycle model is found as \SI{0.09}{m}. This corresponds to \SI{9}{\percent} of the wheelbase of the used bicycle model. The agent trained with the alt(3) shows similar performance.

\subsection{Bicycle Ride along the Benchmark Path in Detail}
\label{Bicycle Ride along the Benchmark Path in Detail}
In this section, a simulated bicycle ride along the benchmark path is analysed in more detail, with the \texttt{agent 1} trained with the proposed settings doing path following and stabilisation of the bicycle model. The forward velocity is initialised at \SI{4}{ms^{-1}}.

\begin{figure}[b]
	\centering
	\includegraphics[scale=1]{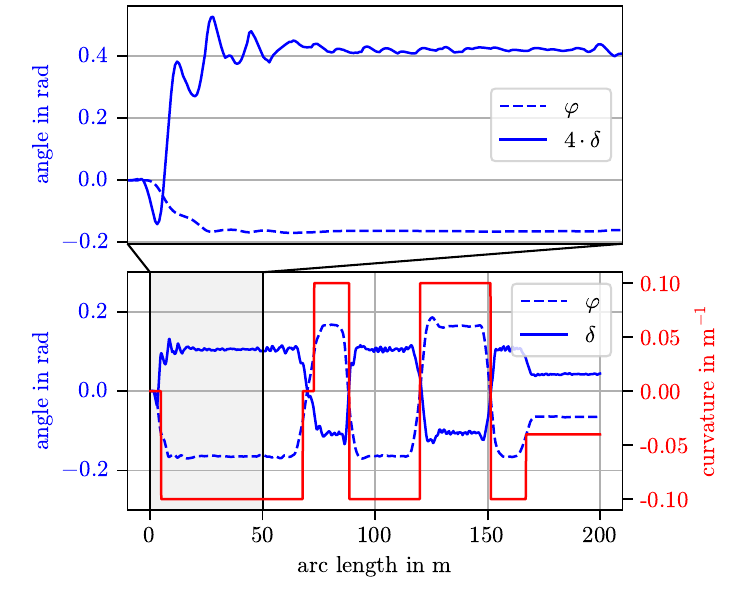}
	\caption{Roll angle $\varphi$, steering angle $\delta$, and curvature $\kappa$ of the path plotted over the first \SI{200}{m} of the benchmark path. \texttt{Agent1} trained with the proposed settings is used to do path following and stabilisation of the bicycle model. The bicycle model travels at \SI{4}{ms^{-1}}.}
	\label{fig:R_propertiesOverArcLength}
\end{figure}

\begin{figure}
	\centering
	\makebox[\textwidth][c]{\includegraphics[scale=1]{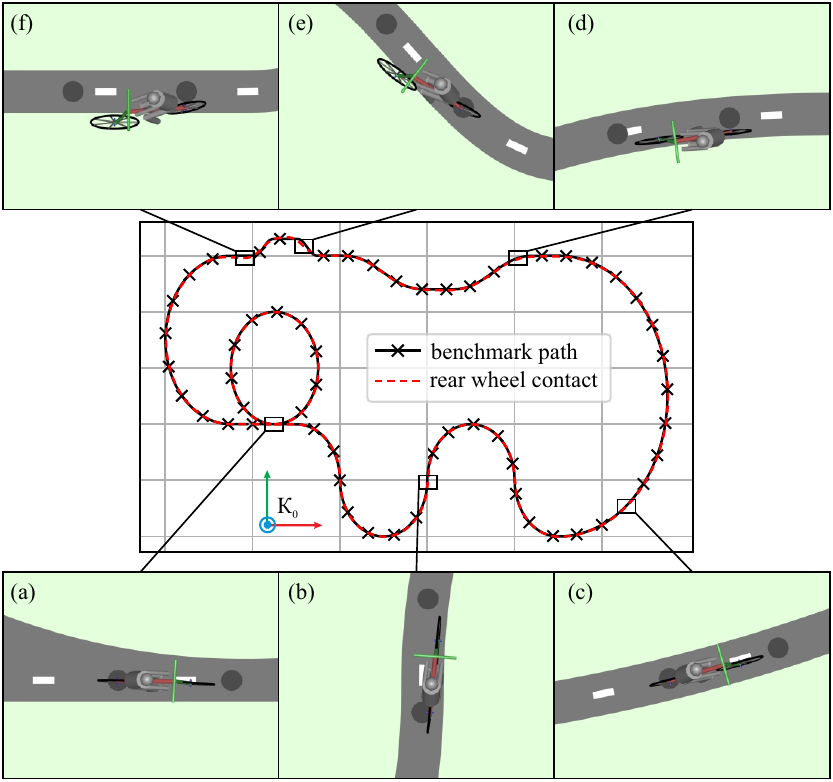}}
	\caption{Rear wheel contact point \Pp\ tracked when the \texttt{agent 1} trained with the proposed settings is used to do path following and stabilisation of the bicycle model along the benchmark path. Six different frames (a)-(f) from the visualisation of the multibody simulation are shown at specific points of the simulated bicycle ride. The circles shown on the street illustrate the points $\mathrm{P_{P0}}, \dots, \mathrm{P_{P4}}$ used to get the preview information of the path, see Figure \ref{fig:RL_pathPreview}.}
	\label{fig:R_simulatedBicycleRide}
\end{figure}

In Figure \ref{fig:R_propertiesOverArcLength}, the roll angle $\varphi$, the steering angle $\delta$, and the (negated) curvature $\kappa$ of the benchmark path are plotted over the first \SI{200}{m} of the benchmark path. It is seen that the roll angle $\varphi$ of the bicycle model is proportional to the curvature $\kappa$ of the path. In the zoomed-in area, it is seen that the steering angle $\delta$ is permanently adjusted by the agent and that $\delta$ gets negative prior to the left curved circle. Such a counter-steering action is necessary to steer the bicycle in the desired direction \cite{Astrom2005}, refer to Section \ref{Resetting the Environment}. Figure \ref{fig:R_simulatedBicycleRide}, frame (a) also shows this counter-steering action done before the bicycle follows the full circle. With the counter-steering, the roll angle $\varphi$ gets negative. Along the full circle, a negative roll angle $\varphi$, induced by the prior counter-steering, is kept. The agent sets the steering angle $\delta$ positive, keeping the bicycle model upright by steering into the direction of the undesired fall \cite{Kooijman2011}, see also introductory Section \ref{Introduction}, and doing path following along the circle. Frame (b) refers to a point where the curvature $\kappa$ changes the sign. Frame (c) shows the bicycle travelling along the flat curve of the path. In frame (d) the bicycle enters the first polynomial transfer. Frames (e) and (f) show the bicycle travelling along the so-called hard lane change. It is seen that the hard lane change is challenging to follow with the bicycle travelling at \SI{4}{ms^{-1}}.



\subsection{Explanation for the Predictions of the Agent}
\label{Explanation for the Predictions of the Agent}
In Section \ref{Bicycle Ride along the Benchmark Path in Detail}, a simulated bicycle ride along the benchmark path using the \texttt{agent 1} trained with the proposed settings is described in detail. During the slalom part of the benchmark path, the state vector $\sv$ is saved every time the agent interacts with the environment so that observations are gained that equally represent the bicycle model doing path following along left and right curves. Using these observations, the predictions of the agent are explained using the SHapley Additive exPlanations (SHAP) values. SHAP values are a measure for the feature importance, introduced in the work \cite{Lundberg2017}. The features of a neural network are the elements in the input vector, being the elements of the state vector $\sv$. The permutation explainer, implemented in the eponymous Python package SHAP\footnote{\url{https://github.com/shap/shap} (used version: 0.45.0)} is used to compute the SHAP values. For each individual observation, being a saved state vector $\sv$, every feature can be assigned a SHAP value. A positive SHAP value of a feature in an observation means that the feature positively influences the prediction of the agent in that observation, while a negative SHAP value means that the feature negatively influences the prediction. Remember that the prediction of the agent is the set value $\delta_\mathrm{set}$ for the steering angle.

\begin{figure}[h]
	\centering
	\includegraphics[scale=1]{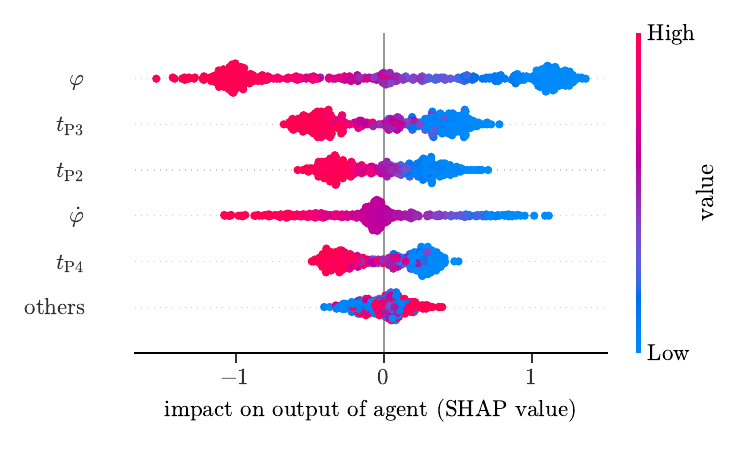}
	\caption{Impact of the features on the prediction of a trained agent measured by the SHAP values. Each dot represents a single observation. The vertical alignment of the dot indicates whether the output of the agent is positively or negatively influenced by the corresponding feature in the observation. The colour of the dot indicates whether the value of this feature is high or low in the observation.}
	\label{fig:R_shapValues}
\end{figure}

In Figure \ref{fig:R_shapValues}, the importance of the features measured by the SHAP values are shown. The features are ordered by their mean absolute SHAP value. The roll angle $\varphi$ is the feature having the greatest impact on the predictions of the agent. In the observations with a positive roll angle $\varphi$, the roll angle $\varphi$ contributes negatively to the prediction of the agent. Thus, a positive roll angle $\varphi$ pulls the set value $\delta_\mathrm{set}$ for the steering angle down and vice versa. The entry $t_\mathrm{P3}$ of the preview vector $\tv$ is the second most important feature based on the mean absolute SHAP values. Observations with a positive value for $t_\mathrm{P3}$ negatively impact the prediction of the agent. The same applies to $t_\mathrm{P2}$, but in the observations this feature does not contribute that much to the output of the agent as $t_\mathrm{P3}$. There are observations, in which the roll angular velocity $\dot{\varphi}$ has a relatively high impact on the output of the agent. However, in most of the predictions, the roll angular velocity $\dot{\varphi}$ is near to zero, not influencing the set value $\delta_\mathrm{set}$ for the steering angle. The mean absolute SHAP value of $t_\mathrm{P4}$, which represents the preview point that is most ahead of the bicycle, is smaller than that of the roll angular velocity $\dot{\varphi}$. The other features are taken together as \textit{others}, as they do not influence the prediction of the agent as much as the other features do, according to the SHAP values. The fact that $x_\mathrm{\Psi}$ and $y_\mathrm{\Psi}$ describing the yaw of the bicycle are part of \textit{others} is in line with the finding that the yaw representation does not influence the learning process, see Section \ref{Learning Process}, since the yaw is not considered to a high degree by the agent.

\section{Discussion}
\label{Discussion}
This section starts by discussing the different investigated settings for the RL framework. Afterwards, the results shown for the performance of the deployed agents are debated. Finally, the explanation for the predictions of the agent given by the SHAP values are compared to already known mechanisms for controlling a bicycle.

In Section \ref{Learning Process}, it is shown that using a constant preview distance $\Delta s$ or including the part $\rho_\mathrm{\varphi}$ in the reward lowers the learning performance. On the former, we find that scaling $\Delta s$ with the forward velocity reduces the learning volume of the agent. The learnt policy must, because of using the Whipple bicycle, take the forward velocity into account to stabilise the bicycle laterally. The scaling of the preview distance eliminates another speed-dependence of the policy. What is striking at first is that the speed-dependence of the policy is not represented by the SHAP values, see Figure \ref{fig:R_shapValues}, since the front wheel rotation angular velocity $\dot{\theta}_\mathrm{F}$ is classified into \textit{others}. The reason for this is not the lack of speed-dependence in the policy, but the way the SHAP values are computed. In the simulated bicycle ride used to obtain the observations for the explanation, the bicycle travels at a constant forward velocity of \SI{4}{ms^{-1}}, see Section \ref{Explanation for the Predictions of the Agent}. Thus, the permutation explainer iterates over almost constant values for the front wheel rotation angular velocity $\dot{\theta}_\mathrm{F}$ with the prediction of the agent not changing within this iteration. Regarding the reward design, we discuss whether including the roll angle $\varphi$ in a more sophisticated way than using the deviation from zero, see Equation \eqref{eq:rhoPhi}, would make the consideration of the part $\rho_\mathrm{\varphi}$ beneficial for the learning progress, i.e. using the deviation from an optimal roll angle that must be computed for the current position of the bicycle. We also think of using the roll angular velocity $\dot{\varphi}$ and a threshold $\eta_\mathrm{\dot{\varphi}}$ instead of the roll angle $\varphi$ and the roll angle threshold $\eta_\mathrm{\varphi}$ in Equation \eqref{eq:rhoPhi}. Initial tests with $\chi_1=\chi_2=0.5$ show that with these changes the learning improves in comparison to that with the alt(5) settings, but is still worse than when only using $\rho_\mathrm{y}$.

In Section \ref{Performance of the Deployed Agents}, it is shown that all agents trained with the proposed and alt(1-4) settings succeed along the type of paths that is used during the learning process. However, individual agents fail along the benchmark path at the lower end of the investigated range for the forward velocity. To explain this, we have a look at the bicycle rides along the benchmark path and find that the individual agents fail at two specific segments of the benchmark path, namely the full circle and the hard lane change. We suspect that these two segments are not part of the learning process and therefore cause difficulties, which we want to demonstrate in the following. Due to the geometric properties of the hard lane change, see Figure \ref{fig:RL_pathTypes}, the hard lane change does not occur in the paths during learning, see Table \ref{tab:LP_initialisationOfCurriculumParameters}. To estimate the probability $P$ of a full circle being part of a single path during learning, it is assumed that for the bicycle travelling at \SI{2}{ms^{-1}} the path is \SI{120}{m} long, resulting from \steps{max} that is specified in Section \ref{Learning Process with Curriculum Learning}. In the first step, $10^4$ paths are generated at random using the reset procedure described in Section \ref{Resetting the Environment} and the initialisation intervals from the last part of the curriculum. Using the expected value for the uniformly distributed angle $\Phi$ of a circular element, a full circle is made out of approximately five consecutive circular elements with the same curvature. In the randomly generated paths, $20$ paths have five circular elements in a row with the same curvature, resulting in $P=20/10^4=0.002$ for the probability of a full circle. The two segments are therefore most likely neither part of the learning process nor of the validation process, which, in combination with a low forward velocity, leads to individual agents failing. The speed-dependence is due to the increasing difficulty of stabilising the bicycle model as velocity decreases and the real part of the Whipple bicycle's eigenvalues increases \cite{Meijaard2007}. On the one hand, the results for the benchmark path show that segments that are not part of paths used during the learning process can cause difficulties for the deployed agents, in particular at low forward velocities. On the other hand, the results show that most of the deployed agents have induced a policy with which path following and stabilisation of the bicycle model can be done at all investigated forward velocities along a path that includes two segments that the agents statistically have not seen during learning. Looking, for example, at the \texttt{agent 1} trained with the proposed settings, a maximal mean distance of \SI{0.09}{m} is found along the benchmark path at \SI{7}{ms^{-1}}, being \SI{9}{\percent} of the wheelbase of the bicycle model. We feel that this is very accurate, considering that the benchmark path cannot be followed exactly due to the step-like curvature changes along the path and the hard lane change that is challenging to follow especially at higher forward velocities.

In Section \ref{Explanation for the Predictions of the Agent}, SHAP values are used to explain the predictions of an agent. For a better understanding of this explanation, we discuss a scenario in which the bicycle model is travelling on a straight and is to turn left. At first, the elements of the preview vector take on positive values. A combination of three of these elements, with $t_\mathrm{P3}$ being considered the most, see Figure \ref{fig:R_shapValues}, negatively influences the steering angle predicted by the agent. Thus, the agent does not steer the bicycle to the left, but to the right. The agent does counter-steering, which is what we expect the agent to learn, as explained in Section \ref{Resetting the Environment}. With this counter-steering, the contact points \Pp\ and \Qp\ move to the right, the roll angle $\varphi$ gets negative, and the bicycle starts to lean into the left turn. As the agent learns to steer into the direction of the undesired fall, see Section \ref{Introduction}, a negative roll angle $\varphi$ pulls the prediction of the agent up, which means that the agent turns the handlebar to the left, as the bicycle leans to the left. The steering angle $\delta$ gets positive and the leaned bicycle follows the left turn. We find that the agent also considers the roll angular velocity $\dot{\varphi}$, which is suspected, as some bicycle controllers use $\dot{\varphi}$ as controlled variable, see for example the intuitive approach described in \cite{Schwab2008}. Taken together, in the trained agent we find two fundamental mechanisms of controlling bicycles considering the SHAP values, that is, to steer into the direction of the undesired fall and to do counter-steering.

\section{Conclusion}
\label{Conclusion}
Before future research is addressed, the present work is concluded by five key statements:
\begin{itemize}
	\item It is demonstrated that a Reinforcement Learning (RL) approach can be used to do path following with the Whipple bicycle while simultaneously stabilising it laterally. No stabilisation aids are needed. The agent learns to do path following and stabilisation of the bicycle model exclusively by setting the steering angle.
	\item It is shown that the RL approach works for a wide range of forward velocities and that a speed-dependent policy is induced.
	\item Curriculum learning as a training strategy is used to make the RL approach applicable for the wide range of forward velocities and arbitrary paths.
	\item An agent is presented that can be used like a conventional controller, being able to steer the bicycle model along a benchmark path with a maximum mean distance of \SI{0.09}{m} over the entire velocity range from $\SI{2}{ms^{-1}}$ to $\SI{7}{ms^{-1}}$. The benchmark path consists of a full circle, a slalom, curves, two lane changes, and straight elements.
	\item The explanation of the agent's predictions shows that the agent has learnt basic mechanisms for controlling bicycles. This builds a bridge to research in the field of bicycle dynamics and to controllers designed using control engineering.
\end{itemize}
Future research might address the practical realisation of a system where a trained agent does path following and stabilisation of the bicycle while its forward velocity is set by the cyclist. Here, above all, the modelling of the bicycle must be refined in order to facilitate the Sim-to-Real transfer, explained in the work \cite{Zhao2020} for RL in the field of robotics. The adaption of the model concerns not only the bicycle model itself, but also the path on which the bicycle travels, being, for example, uneven. In addition, future work must consider a way of obtaining the preview information of the path. Possible approaches might be camera systems or the usage of mapped cycling routes.

\section*{Acknowledgments}
The computational results presented here have been achieved partly using the LEO HPC infrastructure of the University of Innsbruck.

\bibliographystyle{vancouver} 
\bibliography{literature}

\newpage
\begin{appendices}
\section{Coordinates Mappings for the Whipple Bicycle}
\label{Coordinates Mappings for the Whipple Bicycle}
\begin{figure}[h]
	\centering
	\includegraphics[scale=1]{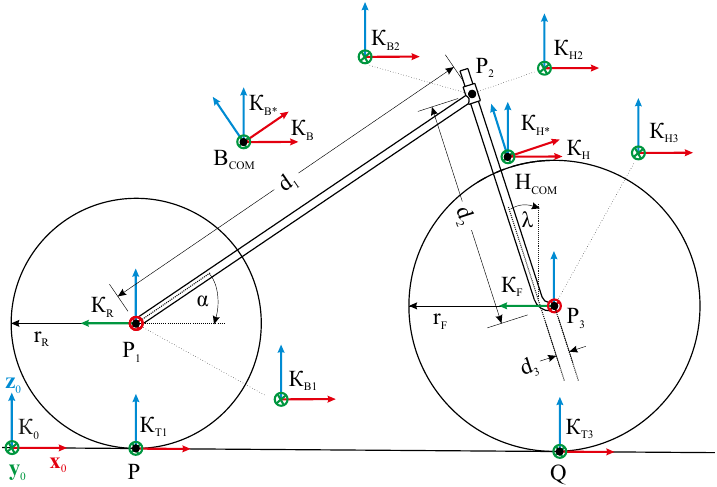}
	\caption{Whipple bicycle shown in the reference configuration, see also Figure \ref{fig:BM_characteristics}, with characteristic geometric parameters and coordinate systems that are needed for the coordinates mappings. Each coordinate system $\mathcal{K}_i$ is described by its origin and unity vectors $\xv_i$, $\yv_i$, and $\zv_i$.}
	\label{fig:Appendix_CharacteristicsDetailed}
\end{figure}

\setcounter{equation}{0}\renewcommand\theequation{\thesection\arabic{equation}}
In this section, it is shown how the output of a multibody simulation of the Whipple bicycle can be mapped to the minimal coordinates of the bicycle model. Afterwards, the mapping from the bicycle's minimal coordinates to the redundant coordinates is described. The relation between the minimal coordinates and the redundant coordinates on both the positional and velocity base of this non-holonomic system is shown in detail. In Table \ref{tab:Appendix_Notation} the notation used in this section is given. The reference frames and geometric properties drawn in Figure \ref{fig:Appendix_CharacteristicsDetailed} are used in the following.

\begin{table}[h]
	\tbl{Notation and spatial transformations used for the coordinates mappings.}{
		\begin{tabular}{c|p{11cm}}
			\toprule
			\textbf{notation} & \textbf{description} \\
			\midrule
			$\Null$ & zero matrix (or zero vector) \\
			$\One$ & unity matrix \\
			$\LUR{\Kframe}{\rv}{\mathrm{P_1}\mathrm{P_2}}$ & vector pointing from the point $\mathrm{P_1}$ to the point $\mathrm{P_2}$ with its coordinates formulated in the \Kframe-frame \\
			$\LU{\Kframe\Mframe}{\Rm}$ & rotation that rotates coordinates of vectors formulated in the \Mframe-frame into the \Kframe-frame: $\LU{\Kframe}{\rv}=\LU{\Kframe\Mframe}{\Rm}\cdot \LU{\Mframe}{\rv}$ \\
			$\LU{\Kframe\Mframe}{\Tm}$ & homogeneous transformation that transforms	coordinates of vectors formulated in the \Mframe-frame into the \Kframe-frame \\
			$\widetilde{\av}$ & skew symmetric matrix of the vector $\av$;	used to calculate the cross product $\cv = \av \times \bv$ as a matrix vector multiplication: $\cv = \widetilde{\av} \cdot \bv$\\
			$\Rotation{\mathrm{e}}{\alpha}$ & rotation by $\alpha$ around the unity vector $\ev$ \\
			\bottomrule
	\end{tabular}}
	\label{tab:Appendix_Notation}
\end{table}

The global \Gframe-frame, the \TIframe-frame, and the \TIIIframe-frame lie in the ground plane. The origin of the latter two frames each lies in the contact point of the wheel and its z-axis points towards the respective wheel hub. The \Rframe-, \Bframe-, \Hframe-, and \Fframe-frame are body-fixed frames of the eponymous bodies with their origin each being located at the COM of the rigid body. The \Rframe- and the \Fframe-frame are defined in a way such that their x-axes are the rotation axes of the wheels. The \Bframe- and the \Hframe-frame can each be translated to two further points on the corresponding rigid body which is indicated with the subscript $\mathrm{1}$ and $\mathrm{2}$ for the rear body and with $\mathrm{2}$ and $\mathrm{3}$ for the handlebar. Two reference frames are marked with a star. They are obtained by rotating the \Bframe-frame by the negated frame angle \alp\ around its y-axis and the \Hframe-frame by the negated steering axis tilt \lam\ around its y-axis. In the following mappings also the two wheel radii \rR\ and \rF, the frame length \dI, the fork length \dII, and the fork offset \dIII\ are used.

\subsection{Mapping of Redundant to Minimal Coordinates using Sensors}
\label{Mapping of Redundant to Minimal Coordinates using Sensors}
To calculate the minimal coordinates, sensors are defined in the interface of the multibody simulation code. One sensor measures the position of the contact point \Pp\ and thus provides the minimal coordinates $x_\Pp$ and $y_\Pp$. Another sensor gives the rotation matrix $\LU{\Gframe\Bframe}{\Rm}$ describing the spatial orientation of the rear body in the global frame. Since the global frame can be turned into the \Bframe-frame by performing a sequence of turns
\begin{equation}
	\LU{\Gframe\Bframe}{\Rm} = \Rotation{z}{\Psi} \cdot \Rotation{x}{\varphi} \cdot \Rotation{y}{\theta_\mathrm{B}} \eqcomma
\end{equation}
the yaw angle $\Psi$ and the roll angle $\varphi$ can be computed using the equations given in \cite{Henderson1977}. The angle $\theta_\mathrm{B}$ is the pitch of the frame and not further needed for the mapping of redundant to minimal coordinates. The steering angle $\delta$, the rear wheel rotation angle $\theta_\mathrm{R}$, and the front wheel rotation angle $\theta_\mathrm{F}$ are output variables of sensors measuring the corresponding revolute joint angles. The roll angular velocity $\dot{\varphi}$ is the component of the rear body angular velocity in the direction of the x-axis of the \TIframe-frame. Thus, it can be written as the inner product
\begin{equation}
	\dot{\varphi} = \langle \LUR{\Gframe}{\xv}{\TIframe}, \angularVelocity{\Gframe}{\Bframe} 
	\rangle \eqcomma
\end{equation}
where $\angularVelocity{\Gframe}{\Bframe}$ is the output quantity of a sensor attached to the rear body. The x-axis $\LUR{\Gframe}{\xv}{\TIframe}$ is known via a homogeneous transformation matrix reading
\begin{equation}
	\LU{\Gframe\TIframe}{\Tm} =
	\underbrace{
		\HT{\Rotation{z}{\Psi}}{\LUR{\Gframe}{\rv}{\Gframe\Pp}}}_{\LU{\Gframe\Cframe}{\Tm}} \cdot
	\HT{\Rotation{x}{\varphi}}{\Null} \eqdot
	\label{eq:T0T1}
\end{equation}
The remaining two minimal coordinates, namely the steering angular velocity $\dot{\delta}$ and the front wheel rotation angular velocity $\dot{\theta}_\mathrm{F}$ are measured by sensors attached to the corresponding revolute joint.

\subsection{Mapping of Minimal to Redundant Coordinates}
\label{Mapping of Minimal to Redundant Coordinates}
In the following, it is shown how the redundant coordinates on positional and velocity base of the four rigid bodies the Whipple bicycle consists of can be computed from the minimal coordinates of the Whipple bicycle.

At first, four homogeneous transformation matrices are defined, representing the position and orientation of the bodies. The pose of the rear body is given by
\begin{equation}
	\LU{\Gframe\Bframe}{\Tm} = \underbrace{\LU{\Gframe\TIframe}{\Tm} \cdot \HT{\Rotation{y}{\theta_\mathrm{B}}}{\LUR{\TIframe}{\rv}{\Pp\PI}}}_{\LU{\Gframe\BIframe}{\Tm}}
	\cdot \HT{\One}{\LUR{\Bframe}{\rv}{\PI\BCOM}} \eqdot
	\label{eq:T0B}
\end{equation}
$\LU{\Gframe\TIframe}{\Tm}$ follows from equation \eqref{eq:T0T1}. The pitch angle $\theta_\mathrm{B}$ satisfies the fact that the rear body can pitch, which means that it can rotate around its y-axis, see Figure \ref{fig:BM_minimalCoordinates}. In \cite{Psiaki1979} it is shown that an analytical solution for $\theta_\mathrm{B}$ given the bicycle minimal coordinates exists. In \cite{Peterson2008} a geometric way to derive a fourth degree polynomial in $\theta_\mathrm{B}$ is introduced. In Appendix \ref{Pitch Angle and Pitch Angular Velocity of the Whipple Bicycle with Rotation Matrices} a convenient way using rotation matrices to get the polynomial in $\theta_\mathrm{B}$ and thus $\theta_\mathrm{B}$ is derived. The rear wheel is represented with
\begin{equation}
	\LU{\Gframe\Rframe}{\Tm}=\LU{\Gframe\BIframe}{\Tm}\cdot
	\HT{\Rotation{z}{\frac{\pi}{2}}}{\Null}\cdot
	\HT{\Rotation{x}{\theta_\mathrm{R}}}{\Null} \eqdot
\end{equation}
The transformation $\LU{\Gframe\BIframe}{\Tm}$ used therein results from Equation \eqref{eq:T0B}. Using Equation \eqref{eq:T0B}, also the handle can be represented as follows:
\begin{equation}
	\LU{\Gframe\Hframe}{\Tm}=\LU{\Gframe\Bframe}{\Tm} \cdot 
	\HT{\Rotation{y}{-\lam}}{\LUR{\Bframe}{\rv}{\BCOM\HCOM}} \cdot
	\HT{\Rotation{z}{\delta}}{\Null} \cdot
	\HT{\Rotation{y}{\lam}}{\Null} \eqdot
	\label{eq:T0H}
\end{equation}
The pose of the front wheel reads
\begin{equation}
	\LU{\Gframe\Fframe}{\Tm}=\LU{\Gframe\Hframe}{\Tm} \cdot
	\HT{\Rotation{z}{\frac{\pi}{2}}}{\LUR{\Hframe}{\rv}{\HCOM\PIII}} \cdot
	\HT{\Rotation{x}{\theta_\mathrm{F}}}{\Null}\eqcomma
	\label{eq:T0F}
\end{equation}
where $\LU{\Gframe\Hframe}{\Tm}$ is taken from Equation \eqref{eq:T0H}.

Furthermore, the translational and angular velocities of the rigid bodies must be computed, using the bicycle's minimal coordinates. The following sequence of computations is followed in the present work:
\begin{enumerate}
	\item angular velocity of \Bframe-frame
	\item angular velocity of \Hframe-frame
	\item angular velocity of \Fframe-frame
	\item translational velocity of \Fframe-frame
	\item translational velocity of \Hframe-frame
	\item translational velocity of \Bframe-frame
	\item translational velocity of \Rframe-frame
	\item angular velocity of \Rframe-frame
\end{enumerate}
It is started with the angular velocity of the rear body, that reads
\begin{equation}
	\angularVelocity{\Gframe}{\Bframe} = \vr{0}{0}{\dot{\Psi}} + \LU{\Gframe\TIframe}{\Rm}
	\vr{\dot{\varphi}}{\dot{\theta}_\mathrm{B}}{0} \eqdot
\end{equation}
In Appendix \ref{Pitch Angle and Pitch Angular Velocity of the Whipple Bicycle with Rotation Matrices} it is described how the pitch angular velocity $\dot{\theta}_\mathrm{B}$ can be computed using rotation matrices. The yaw angular velocity $\dot{\Psi}$ results from
\begin{equation}
	\dot{\Psi} = \frac{1}{w} \sin{(\mu_1-\Psi)} v \eqcomma
\end{equation}
where the forward velocity $v$ of the bicycle is computed using the front wheel rotation angular velocity $\dot{\theta}_\mathrm{F}$ and the radius \rF\ of the front wheel: $v=\rF \cdot \dot{\theta}_\mathrm{F}$. The wheelbase $w$ of the bicycle is computed using Equation \eqref{eq:w}. The angle $\mu_1$ describes the direction of motion of the front wheel contact point \Qp. Since the rotation matrix $\LU{\Gframe\Fframe}{\Rm}$ is already known from Equation \eqref{eq:T0F}, this angle can be calculated by taking
\begin{equation}
	\LU{\Gframe\Fframe}{\Rm} = \Rotation{z}{\mu_1} \cdot \Rotation{x}{\mu_2} \cdot \Rotation{y}{\mu_3}
	\label{eq:R0F}
\end{equation}
into account, using the formulas shown in \cite{Henderson1977}. The angles $\mu_2$ and $\mu_3$ are not needed. The angular velocity of the handle is computed using the relative angular velocity between the rear body and the handle which results exclusively from the revolute joint connecting the rear body and the handle. It follows 
\begin{equation}
	\angularVelocity{\Bframe}{\Bframe\Hframe} = \Rotation{y}{-\lam}\vr{0}{0}{\dot{\delta}} \eqcomma
\end{equation}
with \lam\ being the steering axis tilt. Finally, for the handle applies
\begin{equation}
	\angularVelocity{\Gframe}{\Hframe} = \LU{\Gframe\Bframe}{\Rm} \left(\angularVelocity{\Bframe}{\Bframe} + \angularVelocity{\Bframe}{\Bframe\Hframe}\right) \eqdot
\end{equation}
The same procedure is used to get the angular velocity of the front wheel
\begin{equation}
	\angularVelocity{\Gframe}{\Fframe} = \LU{\Gframe\Hframe}{\Rm} \left(\angularVelocity{\Hframe}{\Hframe} + \angularVelocity{\Hframe}{\Hframe\Fframe}\right) \eqcomma
\end{equation}
where the relative angular velocity between the handle and the wheel results from the front wheel hub and reads
\begin{equation}
	\angularVelocity{\Hframe}{\Hframe\Fframe} = \vtretp{0}{\dot{\theta}_\mathrm{F}}{0}\tp \eqdot
\end{equation}
The translational velocity of the front wheel follows with Euler's first theorem for kinematics to
\begin{equation}
	\LUR{\Gframe}{\vv}{\Fframe} = \LUR{\Gframe}{\vv}{\HIIIframe} =
	\LUR{\Gframe}{\vv}{\TIIIframe} + \angularVelocitySkew{\Gframe}{\TIIIframe} \LUR{\Gframe}{\rv}{\Qp\PIII} \eqdot
	\label{eq:v0F}
\end{equation}
The therein used velocity of the front wheel contact point \Qp, that is $\LUR{\Gframe}{\vv}{\TIIIframe}$, is computed using the angle $\mu_1$ and the bicycle's forward velocity $v$.
\begin{equation}
	\LUR{\Gframe}{\vv}{\TIIIframe}=v\vr{\cos{\mu_1}}{\sin{\mu_1}}{0}
\end{equation}
Remember that $v$ is computed using the front wheel rotation angular velocity $\dot{\theta}_\mathrm{F}$. The angular velocity of the \TIIIframe-frame needed for Equation \eqref{eq:v0F} results from that of the front wheel, but without its rotation angular velocity:
\begin{equation}
	\angularVelocity{\Gframe}{\TIIIframe}= \LU{\Gframe\TIIIframe}{\Rm} \left(\LU{\TIIIframe\Gframe}{\Rm} \angularVelocity{\Gframe}{\Fframe} + \vr{0}{-\dot{\theta}_\mathrm{F}}{0} \right) \eqdot
\end{equation}
Now, $\LUR{\Gframe}{\vv}{\Fframe}$ can be computed and it is continued with the velocity of the handle, reading
\begin{equation}
	\LUR{\Gframe}{\vv}{\Hframe} = \LUR{\Gframe}{\vv}{\HIIIframe} + \angularVelocitySkew{\Gframe}{\Hframe} \LUR{\Gframe}{\rv}{\PIII\HCOM} \eqdot
\end{equation}
With the translational velocity of the bicycle's head
\begin{equation}
	\LUR{\Gframe}{\vv}{\BIIframe} = \LUR{\Gframe}{\vv}{\HIIframe} = 
	\LUR{\Gframe}{\vv}{\Hframe} + \angularVelocitySkew{\Gframe}{\Hframe}\LUR{\Gframe}{\rv}{\HCOM\PII} \eqcomma
\end{equation}
the one of the rear body follows to
\begin{equation}
	\LUR{\Gframe}{\vv}{\Bframe} = \LUR{\Gframe}{\vv}{\BIIframe} + \angularVelocitySkew{\Gframe}{\Bframe} \LUR{\Gframe}{\rv}{\PII\BCOM} \eqdot
\end{equation}
The translational velocity of the rear wheel reads
\begin{equation}
	\LUR{\Gframe}{\vv}{\Rframe} = \LUR{\Gframe}{\vv}{\BIframe} =
	\LUR{\Gframe}{\vv}{\Bframe} + \angularVelocity{\Gframe}{\Bframe}\LUR{\Gframe}{\rv}{\BCOM\PI} \eqdot
\end{equation}
Finally, the angular velocity of the rear wheel can be expressed by
\begin{equation}
	\angularVelocity{\Gframe}{\Rframe} = \LU{\Gframe\Bframe}{\Rm}\left(\angularVelocity{\Bframe}{\Bframe} + \angularVelocity{\Bframe}{\Bframe\Rframe}\right) \eqdot
\end{equation}
The relative angular velocity between the rear body and the rear wheel results from the rear wheel hub. It is denoted as
\begin{equation}
	\angularVelocity{\Bframe}{\Bframe\Rframe} = \vtretp{0}{\dot{\theta}_R}{0}\tp \eqdot
\end{equation}
Since the therein used rear wheel rotation angular velocity is not an element of the state vector, see Equation \eqref{eq:qv}, it must be computed. Using the angular velocity of the \TIframe-frame that reads
\begin{equation}
	\angularVelocity{\TIframe}{\TIframe}=\LU{\TIframe\Gframe}{\Rm} \angularVelocity{\Gframe}{\Bframe} + \vr{0}{-\dot{\theta}_\mathrm{B}}{0} \eqcomma
\end{equation} 
the translational velocity of it in its own frame can be computed as shown below.
\begin{equation}
	\LUR{\TIframe}{\vv}{\TIframe} = \LUR{\TIframe}{\vv}{\Rframe} + \angularVelocitySkew{\TIframe}{\TIframe}\LUR{\TIframe}{\rv}{\PI\Pp}
\end{equation}
Only the x-component of this vector is different from zero, which results from the fact that the wheels are slip free to the side. Consequently, the absolute velocity of the rear contact point \Pp\ reads $v_\TIframe= \vtretp{1}{0}{0}\cdot \LUR{\TIframe}{\vv}{\TIframe}$. The velocity $v_\TIframe$ only results from the rear wheel rotational angular velocity $\dot{\theta}_\mathrm{R}$. Thus, the rear wheel rotation angular velocity can be computed with
\begin{equation}
	\dot{\theta}_\mathrm{R} = \frac{v_\TIframe}{\rR} \eqcomma
\end{equation}
where the radius \rF\ of the rear wheel is used. This completes the mapping from the minimal to the redundant coordinates.

\section{Threshold Value for the Steering Angle of the Whipple Bicycle and Step Response of the Steering Angle}
\label{Threshold Value for the Steering Angle of the Whipple Bicycle and Step Response of the Steering Angle}
\setcounter{equation}{0}\renewcommand\theequation{\thesection\arabic{equation}}
In this section, a threshold value for the steering angle $\delta$ of the Whipple bicycle is defined. The threshold value is used to define a range in which the set value $\delta_\mathrm{set}$ for the steering angle the agent can output must be. With this threshold value, the step response of the steering angle $\delta$ is shown, which may be used to design controllers for the bicycle model, such as a PD controller mapping a set value $\delta_\mathrm{set}$ for the steering angle to a steering torque.

As a result of the tilted steering axis and the fork offset of the Whipple bicycle, the direction of motion of the front wheel contact point \Qp\ relative to the \TIframe-frame is not equal to the steering angle $\delta$ of the Whipple bicycle, see Table \ref{tab:Appendix_DirectionOfMotionOfFrontWheelContactPoint}. In the table it can be seen that the direction of motion of the front wheel contact point, given by the angle $\mu_1$, deviates significantly from the steering angle $\delta$, especially in configurations of the Whipple bicycle where $|\varphi| >> 0$ applies. The angle $\mu_1$ is computed using Equation \eqref{eq:R0F}. Since the magnitude of the roll angle of the bicycle model in the present work should not exceed the threshold $\eta_\mathrm{\varphi}=\ang{45}$, see Section \ref{Reward}, and the direction of motion of the front wheel contact point \Qp\ must not exceed \ang{\pm 90}, a limit for the steering angle $\delta$ can be defined. It is assumed that limiting the steering angle $\delta$ to an value of \ang{\pm 70} is sufficient (although $|\mu_1|$ slightly exceeds \ang{90} when the bicycle is rolled by $|\varphi|=\ang{45}$). Consequently, the set value $\delta_\mathrm{set}$ the agent can output for the steering angle must be in the interval of \ang{-70} to \ang{70}.

\begin{table}[h]
	\tbl{Tabular representation of the dependency between the roll angle $\varphi$, the steering angle $\delta$, and the direction of motion $\mu_1$ of the front wheel contact point \Qp\ of the Whipple bicycle. The minimal coordinates except of $\varphi$ and $\delta$ of the Whipple bicycle are zeroed.}{
	\begin{tabular}{c|c|c}
		\toprule
		\textbf{roll angle $\varphi$} & \textbf{steering angle $\delta$} & \textbf{direction of motion of \Qp\ given by $\mu_1$} \\
		\midrule
		\ang{0} & \ang{0} & \ang{0} \\
		\ang{0} & \ang{60} & \ang{58.8} \\
		\ang{0} & \ang{70} & \ang{69.11} \\
		\midrule
		\ang{-45} & \ang{0} & \ang{0} \\
		\ang{-45} & \ang{60} & \ang{81.66} \\
		\ang{-45} & \ang{70} & \ang{92.16} \\
		\midrule
		\ang{45} & \ang{-0} & \ang{0} \\
		\ang{45} & \ang{-60} & \ang{-81.66} \\
		\ang{45} & \ang{-70} & \ang{-92.16} \\
		\bottomrule
	\end{tabular}}
	\label{tab:Appendix_DirectionOfMotionOfFrontWheelContactPoint}
\end{table}

\begin{figure}[h]
	\centering
	\includegraphics[scale=1]{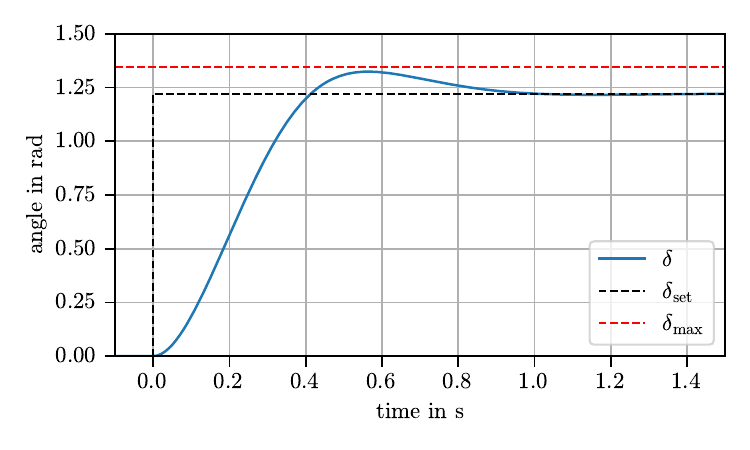}
	\caption{Step response of the steering angle $\delta$ with the bicycle being stationary when the set value $\delta_\mathrm{set}$ for the steering angle follows the function shown. To obtain the step response, the gravitational acceleration is zeroed. The defined maximal permissible value $\delta_\mathrm{max}$ for the step response is also shown.}
	\label{fig:Appendix_stepResponse}
\end{figure}

The step response of the steering angle $\delta$ is used to determine the two parameters $P$ and $D$ of a PD controller that gives the torque $\tau$ that is imprinted between the rear body and the fork of the bicycle model. To get the step response, the stationary bicycle model is simulated when the set value $\delta_\mathrm{set}$ for the steering angle follows
\begin{equation}
	\delta_\mathrm{set}(t)=
	\begin{cases}
		\ang{0} & t < 0\\
		\ang{70} & t \geq 0 \eqdot
	\end{cases}
\end{equation}
For this particular simulation, the gravitational acceleration is zeroed so that the bicycle does not fall over. In Figure \ref{fig:Appendix_stepResponse} the step response of the steering angle $\delta$ is shown. It is defined that the two parameters $P$ and $D$ should be set in a way so that the maximal value of the step response does not overshoot the set value by more than \SI{10}{\%}. By setting $P=\SI{9}{Nm}$ and $D=\SI{1.6}{Nms}$ the specified criterion is fulfilled.

\section{Pitch Angle and Pitch Angular Velocity of the Whipple Bicycle with Rotation Matrices}
\label{Pitch Angle and Pitch Angular Velocity of the Whipple Bicycle with Rotation Matrices}
\setcounter{equation}{0}\renewcommand\theequation{\thesection\arabic{equation}}
In the following, a way to compute the pitch angle $\theta_\mathrm{B}$ of the Whipple bicycle's rear body using rotation matrices is shown, when the minimal coordinates of the bicycle model are given. The notation and reference frames that are used in Appendix \ref{Coordinates Mappings for the Whipple Bicycle} are also used in the following.

It is started by getting the z-axis $\LUR{\Gframe}{\zv}{\TIframe}$ from the \TIframe-frame formulated in the global \Gframe-frame. This vector is the last column of the rotation matrix
\begin{equation}
	\LU{\Gframe\TIframe}{\Rm}=\Rotation{z}{\Psi}\cdot \Rotation{x}{\varphi} \eqdot
\end{equation}
To rotate the \Gframe-frame into the \Bsframe-frame, the rotation matrix
\begin{equation}
	\LU{\Gframe\Bsframe}{\Rm}=\underbrace{\LU{\Gframe\TIframe}{\Rm}\cdot \Rotation{y}{\theta_\mathrm{B}}}_{\LU{\Gframe\Bframe}{\Rm}} \cdot \Rotation{y}{-\alp} 
\end{equation}
is used, with the pitch angle $\theta_\mathrm{B}$ being unknown. The x-axis $\LUR{\Gframe}{\xv}{\Bsframe}$ is the first column of this matrix. The rotation matrix for the \Hsframe-frame follows
\begin{equation}
	\LU{\Gframe\Hsframe}{\Rm}=\LU{\Gframe\Bframe}{\Rm}\cdot \underbrace{\Rotation{y}{-\lam}\cdot \Rotation{z}{\delta}}_{\LU{\Bframe\Hsframe}{\Rm}} \eqdot
\end{equation}
The axes $\LUR{\Gframe}{\xv}{\Hsframe}$, $\LUR{\Gframe}{\yv}{\Hsframe}$, and $\LUR{\Gframe}{\zv}{\Hsframe}$ are the columns of this matrix. As shown in \cite{Peterson2008}, it is possible to calculate a unity vector $\LU{\Gframe}{\hv}$ that points from the front wheel hub in the direction of \Qp\ by performing two steps:
\begin{equation}
	\LU{\Gframe}{\fv}=-\vr{0}{0}{1} + \left(\LUR{\Gframe}{\yv}{\Hsframe}\cdot \vr{0}{0}{1}\right)\eqcomma \qquad 
	\LU{\Gframe}{\hv}=\frac{\LU{\Gframe}{\fv}}{|\fv|} \eqdot
\end{equation}
A vector pointing from \Pp\ to \Qp\ can be computed by
\begin{equation}
	\LUR{\Gframe}{\rv}{\Pp\Qp}= \rR \LUR{\Gframe}{\zv}{\TIframe} + \dI \LUR{\Gframe}{\xv}{\Bsframe} - \dII \LUR{\Gframe}{\zv}{\Hsframe} + \dIII \LUR{\Gframe}{\xv}{\Hsframe} + \rF \LU{\Gframe}{\hv} 
	\eqdot
\end{equation}
Here, \rR, \rF, \dI, \dII, and \dIII\ are bicycle geometries independent of the bicycle's configuration. The wheelbase $w$ of the Whipple bicycle in its current configuration reads
\begin{equation}
	w=|\rv_{\Pp\Qp}| \eqdot
	\label{eq:w}
\end{equation}
Setting the constraint that both wheels each touch the ground at one point leads to a scalar equation reading
\begin{equation}
	\LUR{\Gframe}{\rv}{\Pp\Qp} \vtretp{0}{0}{1}\tp=0 \eqdot
	\label{eq:rPQ_constraint}
\end{equation}
If the bicycle's minimal coordinates are given, the only unknown in this equation is the pitch angle $\theta_\mathrm{B}$. Consequently, a calculation of it with Newton's method with the initial guess $\theta_\mathrm{B}=0$ \cite{Peterson2008} is possible.

A function for the pitch angular velocity $\dot{\theta}_\mathrm{B}$ of the shape $\dot{\theta}_\mathrm{B}(\theta_\mathrm{B}, \varphi, \delta, \dot{\varphi}, \dot{\delta})$ can be gained by deriving Equation \eqref{eq:rPQ_constraint} with respect to time. Thus, also the pitch angular velocity $\dot{\theta}_\mathrm{B}$ is known.
\end{appendices}
\end{document}